\documentclass{article}

\usepackage{soul}
\usepackage{microtype}
\usepackage{graphicx}
\usepackage{subcaption}
\usepackage{booktabs} 
\usepackage{enumitem}
\usepackage{hyperref}
\usepackage{subcaption}

\usepackage{listings}
\usepackage{xcolor}

\usepackage{booktabs}   
\usepackage{tabularx}   
\usepackage{multirow}   
\usepackage{xcolor}     
\usepackage{pifont}     

\usepackage{multirow}

\usepackage[preprint]{icml2026}

\usepackage{amsmath}
\usepackage{amssymb}
\usepackage{mathtools}
\usepackage{amsthm}

\usepackage[capitalize,noabbrev]{cleveref}

\theoremstyle{plain}

\theoremstyle{definition}

\theoremstyle{remark}

\usepackage[textsize=tiny]{todonotes}

\icmltitlerunning{Hindsight Hint Distillation: Scaffolded Reasoning for SWE Agents from CoT-free Answers}
\begin{document}

\twocolumn[
  \icmltitle{Hindsight Hint Distillation: \\ Scaffolded Reasoning for SWE Agents from CoT-free Answers}



  \icmlsetsymbol{equal}{*}

  \begin{icmlauthorlist}
    \icmlauthor{Shengjie Wang}{equal,thu}
    \icmlauthor{Guanghe Li}{equal,thu}
    \icmlauthor{Zonghan Yang}{thu}
    \icmlauthor{Yang Gao}{thu}
  \end{icmlauthorlist}

  \icmlaffiliation{thu}{Tsinghua University}
  \icmlcorrespondingauthor{Yang Gao}{gaoyangiiis@mail.tsinghua.edu.cn}

  \icmlkeywords{Machine Learning, ICML}

  \vskip 0.3in
]



\printAffiliationsAndNotice{\icmlEqualContribution}

\lstset{
    numbers=none
}

\begin{abstract}
Solving complex, long-horizon tasks necessitates strong planning and reasoning capabilities. While datasets with explicit Chain of Thoughts (CoT) would significantly benefit learning, they are costly to obtain. To address this challenge, we propose \emph{Hindsight Hint Distillation} (HHD), which only requires easy-to-obtain, CoT-free $\langle\text{question}, \text{answer}\rangle$ pairs. Inspired by how human teachers leverage student mistakes to provide targeted guidance, HHD synthesizes hindsight hints from the model’s own failed self-rollouts and uses them to scaffold on-policy rollouts that successfully complete the tasks. The model then self-distills these scaffolded trajectories, and generalizes to new problems without hint guidance. Experiments demonstrate that HHD significantly outperforms iterative RFT and trajectory-synthesis baselines, achieving an absolute improvement of 8\% on SWE-bench Verified, whereas all baselines achieve only around 2\%. Notably, the strategy incentives induced by HHD generalize effectively to out-of-distribution tasks, evidenced by the largest gains on SWE-bench Multilingual, despite no training on multilingual data. These results validate that HHD effectively synthesizes expert reasoning from CoT-free data and substantially enhances long-horizon performance.

\end{abstract}

\section{Introduction}

\begin{figure}[t]
  \centering
  \includegraphics[width=0.95\linewidth]{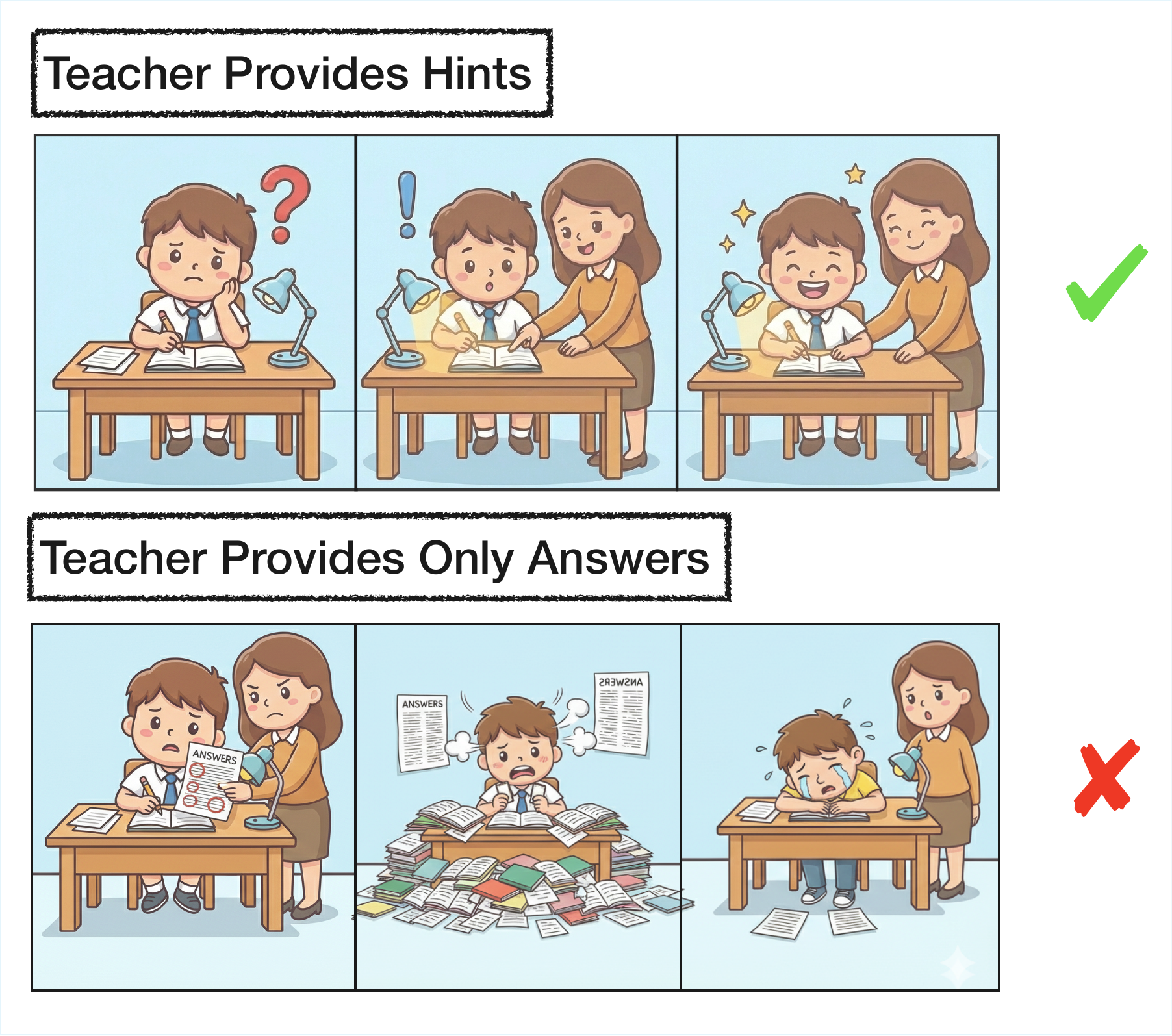}
    \caption{\textbf{Human learning through hints \textit{vs.} rote memorization.} Effective teachers provide targeted hints, rather than direct answers, allowing students to internalize solution strategies, and truly ``understand many'' from one lesson. In contrast, relying solely on answer memorization leads to failure in complex scenarios. Inspired by the hint learning paradigm, we propose \emph{Hindsight Hint Distillation} (HHD) to self-improve using synthetic trajectories scaffolded with summarized hindsight from failed self-rollouts.}
  \label{fig:teaser}
  \vspace{-10pt}
\end{figure}

Large language models (LLMs) have demonstrated strong generalization capabilities across a wide range of tasks, including mathematics, programming, and theorem proving \cite{achiam2023gpt,anthropic_claude_3.5_sonnet_20241022,team2025kimi}. 
A key factor underlying these advances is the use of explicit chain-of-thought (CoT) reasoning, which enables models to decompose complex problems into intermediate steps and thereby improves both reasoning accuracy and robustness \cite{guo2025deepseek, muennighoff2025s1}. 

Motivated by this observation, a substantial body of recent work has focused on acquiring high-quality CoT supervision during the post-train stage~\cite{lai2025survey}.
Although internet-scale repositories provide abundant $\langle\text{question}, \text{answer}\rangle$ data, such as GitHub pull request commits, these sources are predominantly chain-of-thought--free (CoT-free), containing only final outputs or diff patches, without the underlying reasoning processes \cite{jimenez2023swe}. As a result, direct supervision on such datasets fails to teach generalizable problem-solving heuristics.
While manual annotation or synthetic data generation can partially address this limitation, both approaches are prohibitively expensive and difficult to scale, particularly for complex, long-horizon tasks requiring long-horizon reasoning. Self-improvement methods, such as rejection-sampling fine-tuning (RFT) and reinforcement learning (RL), attempt to enhance performance through exploration and exploitation over generated trajectories \cite{shao2024deepseekmath, chen2025minimax}. However, for complex, long-horizon tasks, relying solely on a model’s own exploration to acquire high-quality reasoning remains inherently challenging \cite{yue2025does}. Consequently, how to effectively leverage large-scale CoT-free $\langle\text{question}, \text{answer}\rangle$ data to improve reasoning on complex tasks remains an open problem.

An insightful observation from human learning is that teachers rarely provide complete solution procedures for rote memorization purposes, as illustrated in Figure~\ref{fig:teaser}. Instead, they offer targeted guidance after students make mistakes, enabling students to revisit and re-solve the problems under this guidance and thereby internalize the underlying solution strategies. More importantly, this learning process naturally generalizes to solving similar problems, even without further hints, echoing the Chinese proverb \textit{``learn one thing and understand many''}.

Inspired by this generalizable paradigm of human learning, we propose \emph{Hindsight Hint Distillation} (HHD). By analyzing failed attempts in hindsight to generate targeted hints, HHD scaffolds the model for successful on-policy rollouts. Self-distilling these trajectories enables the model to generalize rather than rote imitation. Under the same training setup on the SWE-Gym dataset, models trained with HHD achieve an 8\% absolute improvement on SWE-bench Verified, whereas other trajectory-synthesis baselines yield at most a 2\% gain. Moreover, despite not incorporating any multilingual training data, HHD-trained models demonstrate substantially stronger generalization on SWE-bench Multilingual compared to the baseline. Our empirical results and detailed case studies demonstrate that training on hint-augmented trajectories significantly enhances the model's generalization capabilities. These results indicate that the HHD framework moves beyond simple rote memorization, instead adopting a learning paradigm akin to human pedagogy: teaching through hints. By leveraging these guided trajectories, the model internalizes underlying logic rather than merely replicating training data.

\section{Related Work}

\subsection{Learning from Guidance for LLMs}

Self-generated trajectory via correction has emerged as a key strategy for addressing the scarcity of high-quality reasoning data in LLM training. Early methods, such as SCLR~\cite{kumar2024training}, applied reinforcement learning (RL) for self-correction but suffered from instability and inconsistency on complex tasks~\cite{olausson2023self}. More recent approaches emphasize structured feedback mechanisms. For instance, Gemini 2.5 Pro~\cite{huang2025gemini, zhou2025solving} utilizes formal bug reports (e.g., Lean 4) to guide iterative reflection–correction cycles, whereas Reflexion~\cite{shinn2023reflexion} employs verbal reinforcement learning to convert environmental feedback into linguistic self-reflections stored in memory.

To overcome the “coverage wall” in generalization~\cite{qin2025decomposing}, recent frameworks focus on synthesizing trajectories that encode diverse error-recovery behaviors. Methods such as Agent-R~\cite{yuan2025agent}, Agent-RLVR~\cite{da2025agent}, and Off-Trajectory Reasoning~\cite{li2025off} leverage environmental rewards and expert-guided corrections~\cite{lauffer2025imitation} to generate traces from off-path failures. By decoupling high-level planning from execution~\cite{erdogan2025plan} and incorporating recursive reasoning~\cite{schroeder2025thread}, these approaches produce synthetic data that enhance long-horizon reasoning and facilitate the transfer of actionable patterns across tasks~\cite{sun2025scaling,anonymous2025shepherd,shi2026r}. However, they predominantly rely on in-episode or step-level reflection. In contrast, our approach summarizes transferable cross-episode hints, avoiding inefficient exploration and thereby improving training efficiency. While QuestA~\cite{li2025questa} exploits partial solutions to reduce task difficulty, it is largely limited to single-turn settings and depends on answer-level supervision with CoT~\cite{li2025questa,wang2025hint}. For complex, long-horizon tasks, obtaining large amounts of CoT-annotated answers remains challenging—a problem that our method directly addresses.

\begin{figure*}[t]
  \centering
  \includegraphics[width=0.95\linewidth]{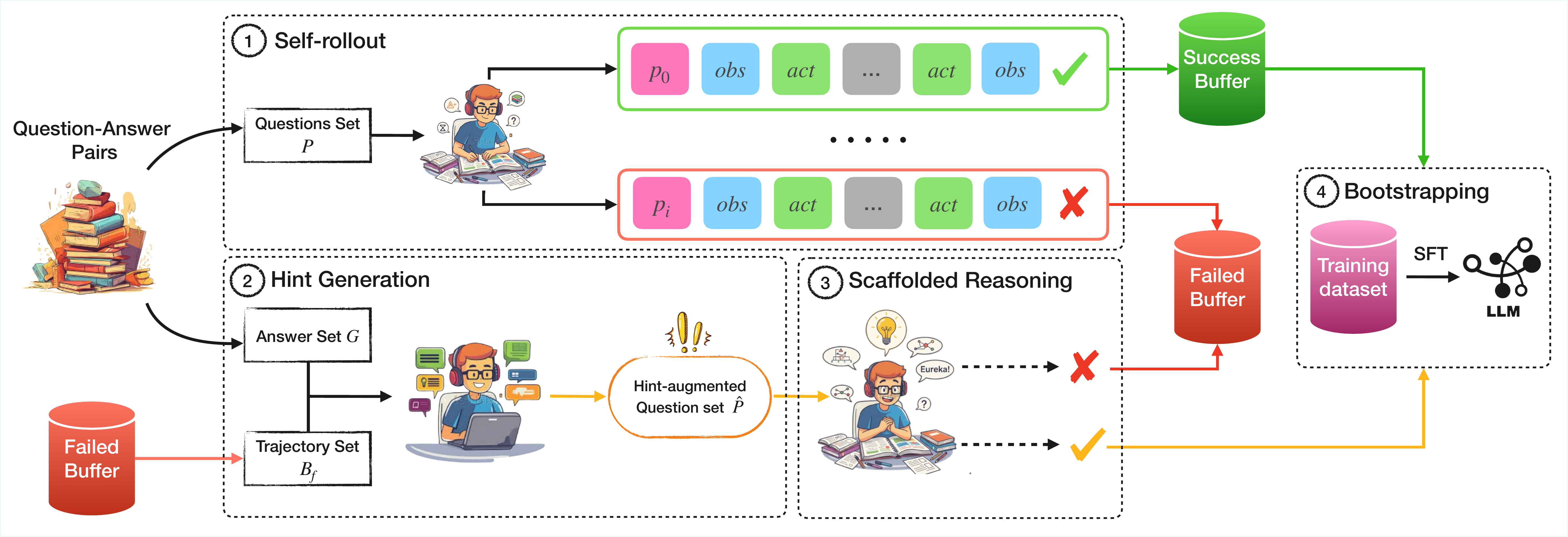}
    \caption{Overview of the Hindsight Hint Distillation (HHD) framework. The pipeline consists of four sequential stages: 
    (1) \emph{Self-rollout}, where the model attempts tasks independently and stores the resulting trajectories in separate success and failure buffers; 
    (2) \emph{Hint Generation}, which leverages failed trajectories ($B_f$) and the ground-truth answer set ($G$) to produce hint-augmented problems ($\hat{P}$); 
    (3) \emph{Scaffolded Reasoning}, where the model re-rolls out solutions conditioned on the generated hints to obtain improved reasoning trajectories; and 
    (4) \emph{Bootstrapping}, which aggregates successful trajectories from both self-rollout and scaffolded reasoning to construct a high-quality training dataset for supervised fine-tuning (SFT).}
  \label{fig:method}
  \vspace{-10pt}
\end{figure*}

\subsection{Coding Agent}

Automated software engineering has evolved from code completion~\cite{chen2021evaluating, li2023starcoder} to autonomous agents that address repository-scale issues~\cite{jimenez2023swe}. Code agent frameworks typically fall into two categories: \textit{agentic scaffolds}~\cite{claude_code, yang2024swe, wang2024openhands}, which enable LLMs to navigate repositories and iteratively edit code via feedback loops, and \textit{workflow-based scaffolds}~\cite{xia2025demystifying, li2025patchpilot, yang2025kimi}, which predefine solving pipelines such as localization, bug repair, and test composition. Recent studies indicate that gains from increasing scaffolding complexity are saturating, with performance now largely determined by the underlying model’s domain expertise rather than the interaction framework~\cite{wei2025swerl}.

Consequently, research has increasingly focused on domain-specific training. To mitigate the scarcity of high-quality trajectories, methods such as SWE-Smith~\cite{yang2025swesmith} and R2E~\cite{jain2024r2e} leverage synthetic data generation and distillation from frontier models to improve sample efficiency. Reinforcement learning (RL) has also gained traction for optimizing long-horizon decision-making. SWE-RL~\cite{wei2025swerl} employs non-expert demonstrations from commit histories, while recent approaches, including Satori-SWE~\cite{zeng2025satori} and DeepSWE~\cite{luo2025deepswe}, combine verifiable execution outcomes with evolutionary strategies or large-scale RL to enhance policy alignment in complex software environments. Kimi-Dev~\cite{yang2025kimi} explores skill transfer between workflow-based and agentic scaffolds, showing in both SFT and RL settings that learning skill priors is crucial. Self-play SWE-RL (SSR)~\cite{wei2025toward} introduces a dual-agent framework in which the model autonomously generates and resolves bugs.

\section{Method}

\subsection{Problem Formulation}

Formally, we consider a large-scale corpus collected from the internet, comprising a question set $P$ and a corresponding answer set $G$. Notably, many real-world data sources are inherently free of CoT, as they typically contain only final outputs, without explicit intermediate steps.

Given each question $p \in P$, the LLM-based agent $\pi$ operates in a sequential decision-making framework. At each time step $t$, the agent receives an observation $o_t$ from the environment and produces an action according to $a_t \sim \pi\bigl(a_t \mid p, \{(o_i, a_i)\}_{i=0}^{t-1}, o_t\bigr)$. After $M$ steps, the trajectory is defined as $\tau = \left\{ p, (o_i, a_i)_{i=0}^{M-1}, o_M \right\}$. The final outcome is evaluated against the ground-truth answer $g \in G$ using either rule-based or model-based evaluators to determine task success.

During the post-training stage, a common paradigm leverages exploration to collect successful trajectories for self-improvement. Specifically, given a question set $P = \{ p_0, \ldots, p_N \}$, the model generates a trajectory buffer $B_i = \{ \tau_i^{(1)}, \ldots, \tau_i^{(K)} \}$ for each problem $p_i \in P$. The model can then improve either by learning solely from successful trajectories (RFT), or by constructing a reward function to rank trajectories and applying reinforcement learning (RL) algorithms to update its parameters.

\subsection{Key Insight}
\label{sec:key_insight}

A fundamental challenge in training agents on long-horizon reasoning tasks is striking the right balance between exploration and exploitation. Existing methods, such as Rejection Sampling Fine-Tuning (RFT) and standard Reinforcement Learning (RL), primarily rely on outcome-based verification. For difficult problems, these approaches often struggle to discover high-quality solution trajectories; even when a correct final answer is produced, the trajectory may result from inefficient trial-and-error or coincidental reasoning. Consequently, the available learning signal is limited, which encourages memorization rather than systematic planning.


Inspired by human learning paradigms illustrated in Figure~\ref{fig:teaser}, we propose Hindsight Hint Distillation (HHD), which replaces unguided exploration with hint-guided search. Rather than sampling trajectories randomly, HHD conditions trajectory generation on high-level hints provided by a teacher, substantially improving exploration efficiency and enabling the discovery of solution paths that are difficult for standard RL methods to reach. Importantly, these hints are derived retrospectively by comparing previous failed attempts with the ground-truth answer, yielding targeted and problem-specific guidance. Conditioning on such hindsight hints allows HHD to generate training trajectories that are both informative and effective, reducing spurious randomness while providing high-quality supervision for learning structured and coherent reasoning patterns.

\subsection{Hindsight Hint Distillation}
\label{sec:hindsight_hint_distillation}

Hindsight Hint Distillation (HHD) is a self-improvement paradigm designed to generate expert-level trajectories by retrospectively analyzing failures. 
As illustrated in Figure \ref{fig:method}, the HHD framework consists of four stages:

\textbf{Self-rollout and Buffer Partitioning}: The process begins with a \textit{Self-rollout} phase, where an LLM-based agent $\pi$ attempts to solve questions from the set $P$. The resulting trajectories are then evaluated against the ground-truth answer set $G$ and partitioned into a \textit{Success Buffer} and a \textit{Failure Buffer}. The latter, denoted as $B_f$, serves as the primary resource for identifying reasoning bottlenecks.

\textbf{Hindsight Hint Generation}: To leverage failed attempts, HHD performs hint extraction by analyzing the gap between the failed trajectories in $B_f$ and the ground-truth answers $G$. This stage produces concise heuristic hints $h$ that highlight overlooked constraints or sub-goals. These hints are then combined with the original questions to form a \textit{Hint-augmented Question Set} $\hat{P}$. A key design choice is that the hints are intentionally brief (typically a single sentence or phrase), ensuring minimal context overhead, preventing the leakage of complete solutions, and encouraging the model to refine its own reasoning rather than simply imitating ground-truth answers. In Section~\ref{sec:experiments}, we demonstrate that longer, more explicit guidance (e.g., Agent-RLVR-Plan) underperforms compared to our concise hint design.

To generate such hints, HHD first applies a rule-based compression to each failed trajectory, as illustrated in Listing~\ref{lst:comp_traj}. This procedure removes most verbose observations while retaining the sequence of model actions along with short, truncated observation snippets, substantially reducing the trajectory context length. In practice, we find that the action sequence alone is sufficient for diagnosing most reasoning failures. 
The compressed trajectory, together with the corresponding ground-truth solution, is then fed into an expert model that produces concise textual hints identifying the underlying causes of failure. Notably, the hint-generation prompt (Listing \ref{lst:hdd_prompt}) is domain-agnostic, allowing the approach to generalize across a wide range of multi-turn reasoning tasks. Concrete examples of generated hints and implementation details are provided in Appendix~\ref{app:impl_hhd}.

\textbf{Hint-Based Scaffolded Reasoning}: Given the hint-augmented problem set $\hat{P}$, we perform scaffolded reasoning by conditioning the agent’s generation on a high-level hint $h$ provided at the start of the trajectory. Formally, actions are sampled as
\begin{equation}
\hat{a}_t \sim \pi\!\left(\hat{a}_t \mid \hat{p}, \{(o_i, a_i)\}_{i=0}^{t-1}, o_t \right),
\end{equation}
where $\hat{p}$ denotes the original problem augmented with the hindsight hint, $\hat{p} = [p,h]$. The concrete format of $\hat{p}$ is provided in Appendix~\ref{app:impl_hhd} (see Listing~\ref{lst:hint_problem}).

Instead of providing additional environment or user feedback during rollout, the hint is supplied only once at the beginning and remains fixed throughout generation. In this way, the hint functions as a global guideline, guiding the agent toward the correct solution without step-by-step external corrections. This design constrains exploration at the strategy level: it prevents the model from repeating previously observed misunderstandings or unproductive trajectories, while allowing the generation of all intermediate reasoning steps autonomously. Consequently, the resulting trajectories remain on-policy, offering consistent and informative learning signals directly sampled from the agent’s own policy, in contrast to off-policy or unguided rollouts.

\textbf{Bootstrapping with Trajectories}: In the final \textit{Bootstrapping} stage, HHD consolidates successful trajectories from both the initial self-rollout and the scaffolded reasoning to construct a high-quality \textit{training dataset}. This procedure is performed iteratively: in each round, the dataset is refreshed, the model is fine-tuned via supervised fine-tuning (SFT) on the updated dataset, and the refined model is used to generate trajectories for the next iteration. This iterative process allows the model to progressively internalize improved reasoning patterns.

Overall, HHD aims to enhance exploration efficiency in long-horizon tasks by leveraging hindsight hints to guide trajectory generation. This approach encourages the agent to focus on promising solution paths that are typically challenging for standard RL methods to discover, while preserving autonomous reasoning and producing informative data for further learning.

\begin{table*}[t]
\centering
\caption{Pass@k performance (\%) under different models and methods. Results are reported on SWE-bench Verified. HHD achieves the best performance across most metrics for both Qwen-72b and GLM-Air models, demonstrating the effectiveness of our approach compared to existing baselines, including Naive RFT, SE-agent-Reflect, and Agent-RLVR variants. HHD-M is an ablation variant of our approach, as described in Section~\ref{sec:ablation}.}
\label{tab:swe-verified}

\setlength{\tabcolsep}{6pt}
\renewcommand{\arraystretch}{1.05}

\begin{tabular*}{\textwidth}{@{\extracolsep{\fill}} l c c c | l c c c}
\toprule
\multicolumn{8}{c}{\textbf{SWE-bench Verified}} \\
\midrule
\multicolumn{4}{c|}{\textbf{Qwen-72b Model}} & 
\multicolumn{4}{c}{\textbf{GLM-Air Model}} \\
\midrule
Method & Pass@1 & Pass@3 & Pass@5 &
Method & Pass@1 & Pass@3 & Pass@5 \\
\midrule
Naive RFT      & 44.0 & 62.2 & 67.2 &
Naive RFT      & 38.0 & 53.0 & 58.8 \\

SE-agent-Reflect      & 43.4 & 64.4 & 69.0 &
SE-agent-Reflect      & 40.0 & 55.2 & 59.8 \\ 

Agent-RLVR-Plan      & 45.6 & 65.4 & \textbf{70.2} &
Agent-RLVR-Plan      & 38.6 & 55.0 & 60.2 \\ 

Agent-RLVR-EF      & 43.4 & 65.0 & 69.0 &
Agent-RLVR-EF      & 39.4 & 54.4 & 59.6 \\ 

Dense Expert Judge   & 45.8 & 65.0 & 68.8 &
Dense Expert Judge      & 39.4 & 55.2 & 59.0 \\ 

\midrule
HHD-M          & 47.2 & 61.8 & 67.8 &
HHD-M          & 41.0 & 54.6 & 59.6 \\

\textbf{HHD (ours)} & \textbf{51.2} & \textbf{67.0} & \textbf{70.2} &
\textbf{HHD (ours)} & \textbf{42.0} & \textbf{55.4} & \textbf{60.8} \\
\bottomrule
\end{tabular*}
\vspace{-10pt}
\end{table*}

\section{Experiments}
\label{sec:experiments}
We evaluate HHD on complex, long-horizon tasks from the SWE-bench benchmark, which typically require nearly 100 turns of interaction. Given a real-world GitHub issue, the model must generate a patch verified by unit tests. HHD shows substantial gains over strong baselines on both SWE-bench Verified and Multilingual, and notably, no additional multilingual training is used while the model generalizes effectively across languages.


\subsection{Setup and Baselines}

\subsubsection{Agent Framework}
We employ \textbf{OpenHands} \cite{wang2024openhands} as the agent framework, which enables agents to leverage tools and interact with the environment. The solving procedure follows a standard software development workflow: (1) \textbf{exploration}, where the agent navigates the repository to locate the relevant files and understand the issue; (2) \textbf{reproduction}, where the agent creates a minimal script to trigger the bug; (3) \textbf{implementation}, involving code modifications and verification against the reproduction script; and (4) \textbf{testing}, where the agent adds edge cases to ensure robustness.
The agent can dynamically backtrack and switch between exploration, coding, and debugging phases until the issue is fully resolved.
Details of the OpenHands framework implementation and the action space can be found in Appendix \ref{app:openhands}.

\begin{figure*}[t]
    \centering
    \begin{subfigure}[t]{0.48\textwidth}
        \centering
        \includegraphics[width=\linewidth]{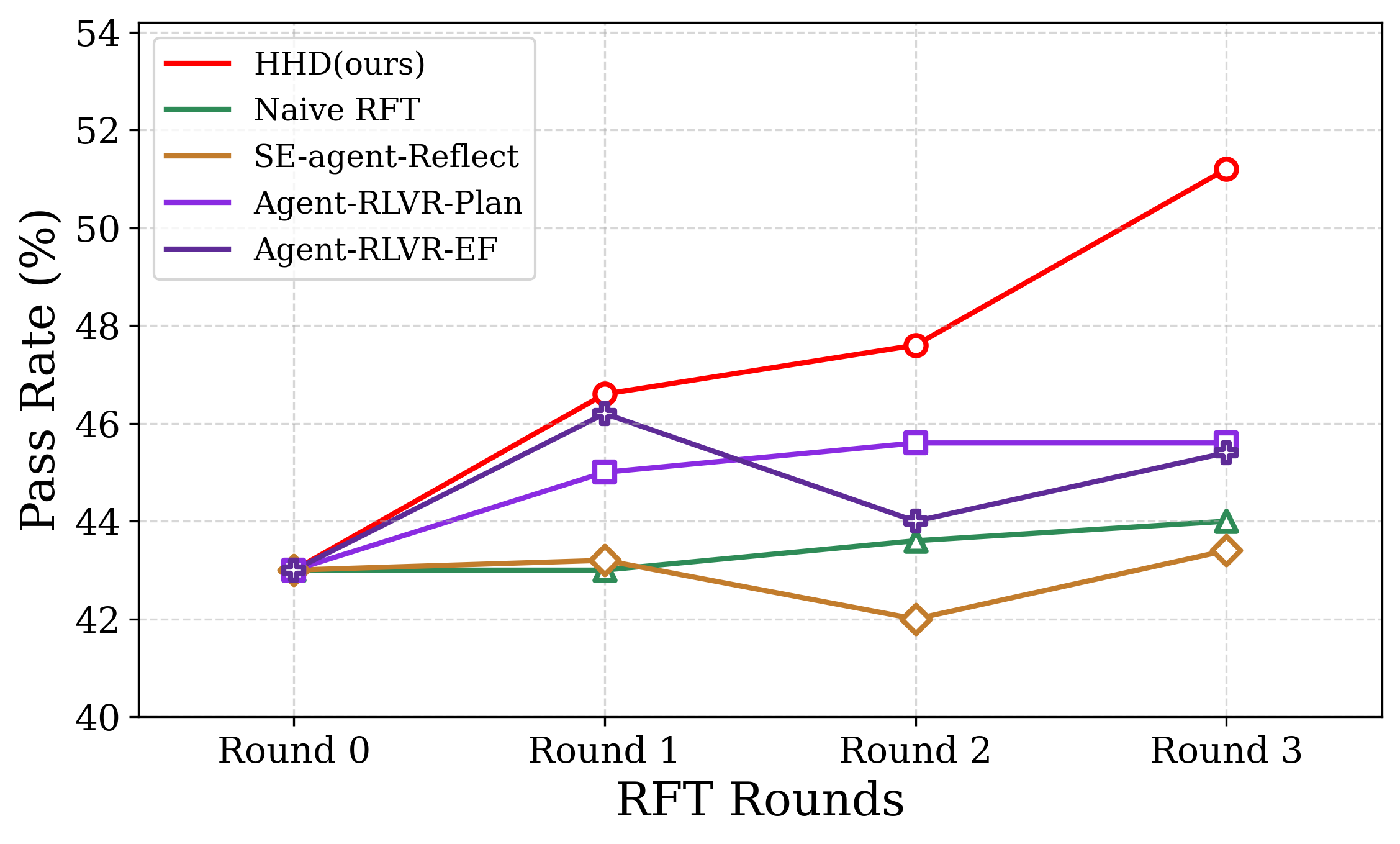}
        \caption{Qwen-72b results}
    \end{subfigure}
    \hfill
    \begin{subfigure}[t]{0.48\textwidth}
        \centering
        \includegraphics[width=\linewidth]{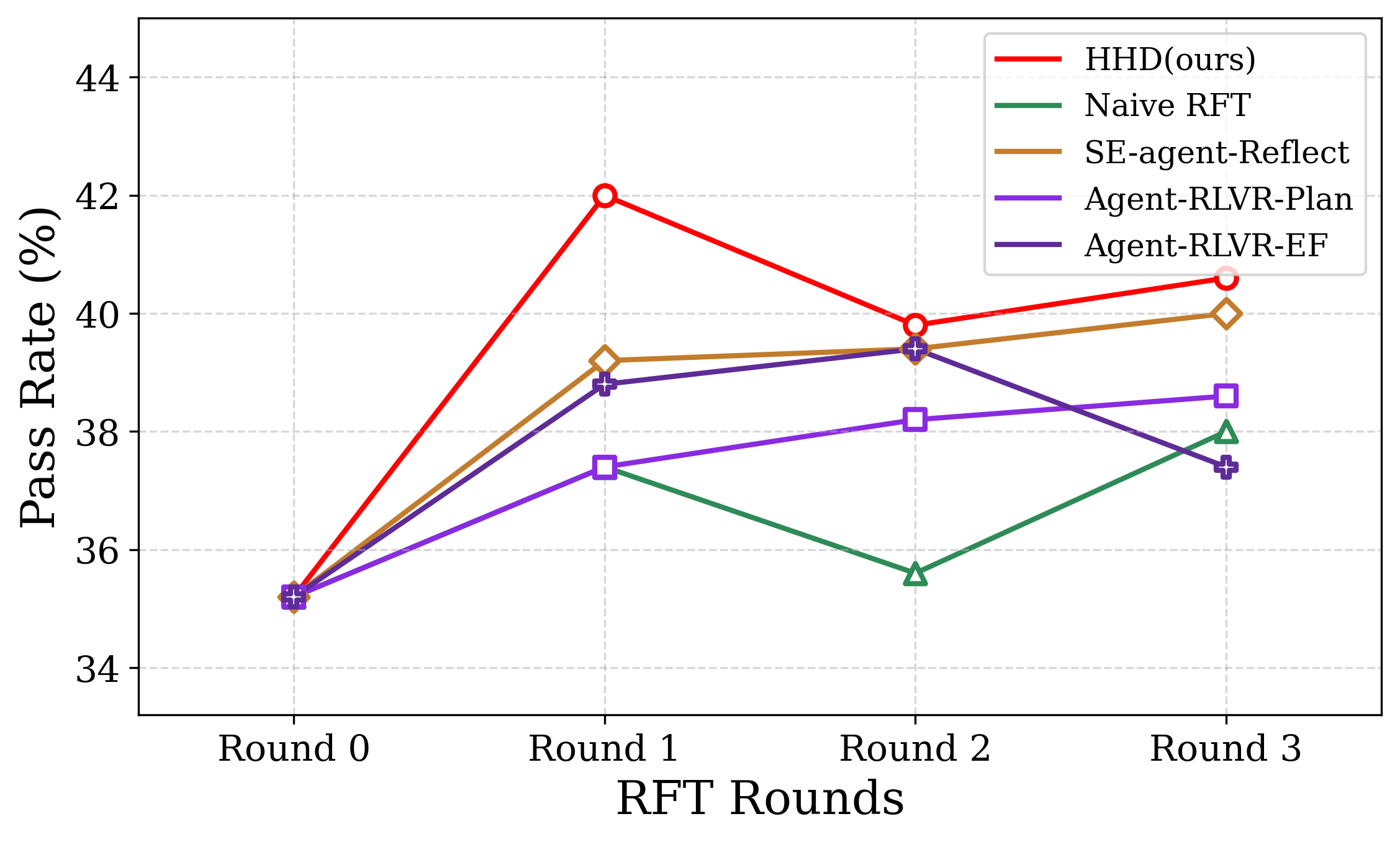}
        \caption{GLM-Air results}
    \end{subfigure}
    \caption{Performance comparison between HHD and baseline methods on Qwen and GLM models. HHD achieves the highest learning efficiency, outperforming naive RFT, SE-agent-Reflect, and Agent-RLVR variants.}
    \label{fig:comparison}
    \vspace{-15pt}
\end{figure*}

\subsubsection{Datasets and Models}
\textbf{Datasets}: We adopt the open-source SWE-Gym dataset for training, which contains 2.4K real-world tasks collected from 11 Python repositories, along with an additional Lite split of 234 instances \cite{pan2024training}.
We evaluate our model on SWE-bench Verified, a human-validated subset of 500 samples from the SWE-bench test set \cite{jimenez2023swe}.
We also evaluate our method on SWE-bench Multilingual \cite{yang2025swesmith}, which consists of 300 curated tasks from 42 GitHub repositories spanning 9 programming languages, including Rust, Java, PHP, Ruby, JavaScript/TypeScript, Go, and C/C++.

\textbf{Models}: For the backbone models, we use Qwen2.5-72B-base \cite{yang2025qwen3} and GLM-4.5-Air-base \cite{zeng2025glm}, representing dense and Mixture-of-Experts (MoE) architectures, respectively. 
To enable agentic behavior within the OpenHands framework, we bootstrap two base models using tool-augmented rollouts generated by the Kimi-K2 model~\cite{team2025kimi} on SWE-Gym dataset. The collected trajectories are filtered via reject-sampling, and the retained successful trajectories are used to fine-tune the base models.
Using this training protocol, Qwen-72B and GLM-Air models achieve pass@1 scores of 43.0 and 35.2, respectively, on SWE-bench Verified. Full training details, including rollout passes, batch size, sequence length, and learning rate schedule, are provided in Appendix~\ref{app:models}.

\subsubsection{Baselines}

We implement the following baselines for comparison.

\textbf{Naive RFT}: For each task in the SWE-Gym dataset, we perform two rollouts per round with a temperature of 1. Only trajectories that successfully reach the correct objective are retained for training, following the standard rejection-sampling fine-tuning (RFT) procedure.

\textbf{SE-agent-Reflect}: Building on Naive RFT, we integrate the reflection mechanism introduced in SE-agent~\cite{lin2025se} to address failed trajectories in each round. The process involves three steps: 
(1) identifying the critical step in the failed trajectory; 
(2) summarizing the solution strategy preceding this step; and 
(3) inserting a prompt at the critical step that incorporates the summary and asks the model to reconsider and propose a new solution. 
Following the original formulation, we adopt the same prompt template, embedding it directly into our implementation. 
As reported in the paper, trajectories generated after reflection exhibit a high likelihood of success in subsequent steps. 
We include these reflected, successful trajectories in the training set.

\textbf{Agent-RLVR}: Agent-RLVR~\cite{da2025agent} is a framework for training agents with RLVR using environment feedback and guidance. From the proposed guidance types, we select two representative ones as baselines: planning guidance and environment feedback (EF), referred to as \textbf{Agent-RLVR-Plan} and \textbf{Agent-RLVR-EF}, respectively. As Agent-RLVR was not originally adapted for coding, we implemented a variant specifically designed for coding tasks. For a fair comparison, we adopt basic SFT instead of the offline DPO used in the original setup.

\textbf{Dense Expert Judge:} Building upon Naive RFT, we adopt a strategy similar to SWE-Search~\cite{antoniades2024swe}: for problems that failed in the initial rollout, we perform a re-rollout employing step-level rejection sampling. At each reasoning or planning step, the model generates multiple candidate actions, from which an expert model selects the optimal one to continue the trajectory. This process effectively functions as a form of guided search via implicit intervention.

For all baselines, the expert model providing guidance is a DeepSeek-V3~\cite{liu2024deepseek} model with access to ground-truth answers. This setup ensures fair and comparable evaluation across methods. Notably, DeepSeek-V3 underperforms our rollout model on SWE-bench Verified, mitigating potential confounding effects from knowledge distillation.

All algorithms follow identical supervised learning settings in each training round. We use a global batch size of 64 and a maximum sequence length of 128k tokens to accommodate the long-context requirements of real-world software engineering tasks. The learning rate is set to $1 \times 10^{-6}$ for the Qwen-72b model and $2 \times 10^{-6}$ for the GLM-Air model, which were found to be the best choice for RFT training. We performed three rounds of RFT, and for each method, the best-performing checkpoint among three candidates was selected for evaluation.

\begin{table*}[t]
\centering
\caption{Pass@k performance (\%) on SWE-bench Multilingual. Without any training on multilingual data, HHD outperforms all baselines, demonstrating its strong cross-lingual generalization.}
\label{tab:swe_multilingual}

\setlength{\tabcolsep}{6pt}
\renewcommand{\arraystretch}{1.05}

\begin{tabular*}{\textwidth}{@{\extracolsep{\fill}} l c c c | l c c c}
\toprule
\multicolumn{8}{c}{\textbf{SWE-bench Multilingual}} \\
\midrule
\multicolumn{4}{c|}{\textbf{Qwen-72b Model}} &
\multicolumn{4}{c}{\textbf{GLM-Air Model}} \\
\midrule
Method & Pass@1 & Pass@3 & Pass@5 &
Method & Pass@1 & Pass@3 & Pass@5 \\
\midrule
Naive RFT & 18.3 & 42.3 & 47.7 &
Naive RFT & 16.7 & 30.3 & 37.3 \\

SE-agent-Reflect      & \textbf{21.0} & 42.3 & 46.6 &
SE-agent-Reflect      & 18.0 & 30.6 & 37.3 \\ 

Agent-RLVR-Plan      & 20.4 & 41.6 & 47.3 &
Agent-RLVR-Plan      & 16.0 & 31.3 & 36.3 \\ 

Agent-RLVR-EF      & 18.3 & 41.6 & 46.7 &
Agent-RLVR-EF       & 17.0 & 30.6 & 34.0 \\ 

Dense Expert Judge    & 18.0 & 41.0 & 46.6 &
Dense Expert Judge      & 17.0 & 32.3 & 38.3 \\   

\midrule
HHD-M          & 17.6 & 41.0 & 45.6 &
HHD-M          & 17.3 & \textbf{35.0} & 39.0 \\

\textbf{HHD} & \textbf{21.0} & \textbf{44.3} & \textbf{48.3} &
\textbf{HHD} & \textbf{18.3} & 33.0 & \textbf{39.0} \\
\bottomrule
\end{tabular*}

\vspace{-15pt}
\end{table*}

\subsection{Comparison}

\textbf{Overall Performance.} Table~\ref{tab:swe-verified} reports the Pass@$k$ results on SWE-bench Verified. HHD consistently achieves state-of-the-art performance across both Qwen and GLM models, outperforming all baselines at every $k$ setting.  
For the Qwen-72b model, HHD reaches 51.2\% Pass@1 and 70.2\% Pass@5, substantially surpassing Naive RFT. Notably, the relative improvement at Pass@1 is larger than at Pass@5, suggesting that HHD improves the average success rate on problems that were previously solved incorrectly on occasion, thereby exhibiting stronger gains in Pass@1. Although the margin narrows at higher $k$, the improvement remains meaningful, as Pass@1 is often the most critical metric in practical deployment scenarios.  
Similarly, for the GLM-Air model, HHD maintains a clear advantage with 42.0\% Pass@1 and 60.8\% Pass@5. The same trend holds, with gains at Pass@3 and Pass@5 being relatively smaller than at Pass@1, reinforcing that HHD primarily enhances first-attempt solution quality rather than diversity-driven performance. 



\textbf{Learning Efficiency}. Figure \ref{fig:comparison} illustrates the evaluation performance throughout the whole RFT process. On Qwen-72B model, HHD exhibits monotonic improvement, maintaining a stable margin over baselines and suggesting a progressive acquisition of effective reasoning behaviors. On GLM-Air model, while the learning curve shows moderate fluctuations, HHD achieves the most rapid performance gains. We attribute this efficiency to HHD's on-policy trajectory generation. In contrast, methods like Agent-RLVR-Plan rely on expert reasoning and suffer from off-policy distribution shift. Furthermore, while SE-agent-Reflect and Agent-RLVR-EF avoid off-policy issues, their reliance on external feedback results in lower learning efficiency compared to HHD’s approach. 

To intuitively illustrate how HHD benefits the model's reasoning capability, we discuss a case study in Appendix~\ref{sec:appendix_detail_of_case_study}. In this case, HHD solves a complex issue with clear reasoning, while the baseline fails to progress, getting stuck in loops.

\subsection{Generalization}

As shown in Table~\ref{tab:swe_multilingual}, HHD demonstrates strong zero-shot generalization on SWE-bench Multilingual, outperforming baseline methods despite not using any multilingual data during training. On the Qwen-72b model, HHD achieves a Pass@1 of 21.0\%, compared to 18.3\% for Naive RFT. Similarly, on the GLM-Air model, Pass@1 improves from 16.7\% to 18.3\% with HHD. We attribute this cross-lingual transferability to the supervision signal provided by HHD: while standard RFT baselines tend to overfit to surface-level Python code patterns, HHD encourages the model to internalize language-agnostic planning strategies. By conditioning generation on high-level hints (e.g., ``isolate the root cause before editing'' or ``verify dependencies''), the model learns abstract reasoning workflows rather than merely reproducing surface-level code syntax, enabling the model to differentiate between issues caused by the local environment and actual codebase bugs, independent of the programming language.

\subsection{Why HHD Works}
\label{sec:analysis}

To understand the factors contributing to HHD's performance gains, we conduct a comprehensive analysis of the generated trajectories from two critical dimensions: \textbf{Rollout Quality} and \textbf{Rollout Efficiency}. 

\begin{figure}[t]
    \centering
    \includegraphics[width=0.95\linewidth]{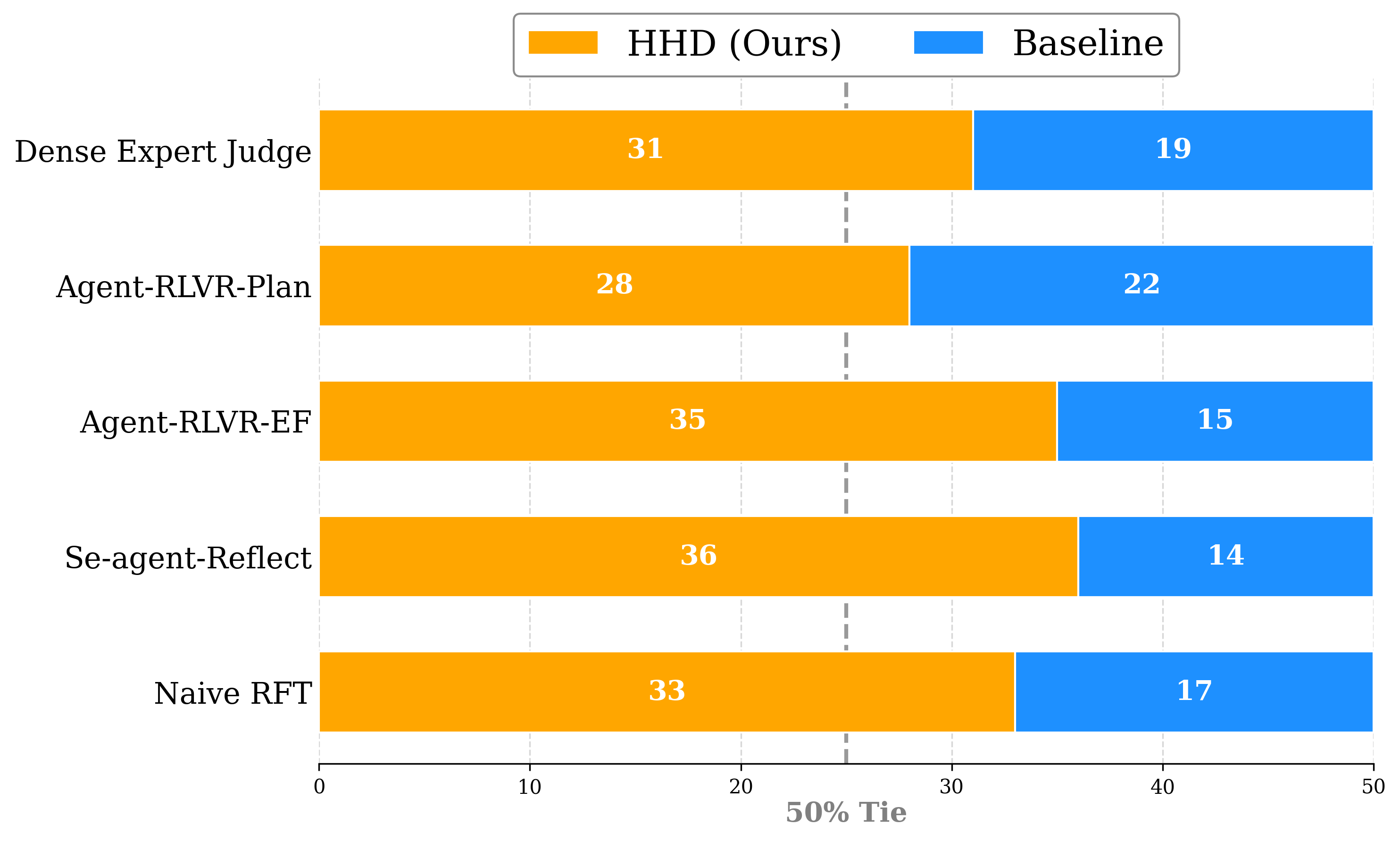}
    \vspace{-0.2cm} 
    \caption{\textbf{Trajectory quality comparison.} 
    To evaluate the quality of reasoning processes, we conducted a pairwise comparison between HHD and baselines. 
    For each baseline, we randomly sampled 50 tasks where \textit{both} HHD and the baseline successfully solved the problem, ensuring a fair comparison of reasoning quality rather than correctness. 
    We then employed Gemini-2.5-Flash as an external judge to determine which trajectory exhibited superior reasoning soundness and coherence (the specific evaluation prompt is detailed in Appendix~\ref{app:judge_prompt}). 
    The results indicate that HHD's trajectories (orange) are consistently preferred over those from baselines (blue), surpassing the 50\% tie threshold across all comparisons.
    }
    \label{fig:quality_comparison}
    \vspace{-10pt}
\end{figure}

\textbf{Rollout Quality Analysis.} Beyond final outcome correctness, the quality of the training data is important. To assess this, we conducted a pairwise comparison of trajectories generated by Qwen-72B across different scaffolding methods, employing Gemini-2.5-Flash as an external judge to evaluate reasoning soundness and coherence (see Appendix~\ref{app:judge_prompt} for detailed prompts). As illustrated in Figure~\ref{fig:quality_comparison}, HHD consistently outperforms all baseline methods, achieving a win rate exceeding 50\% in every comparison. This confirms that HHD solves tasks through superior structured reasoning rather than merely arriving at correct answers via stochastic trial-and-error.

\begin{figure}[t] 
    \centering
     \includegraphics[width=0.85\linewidth]{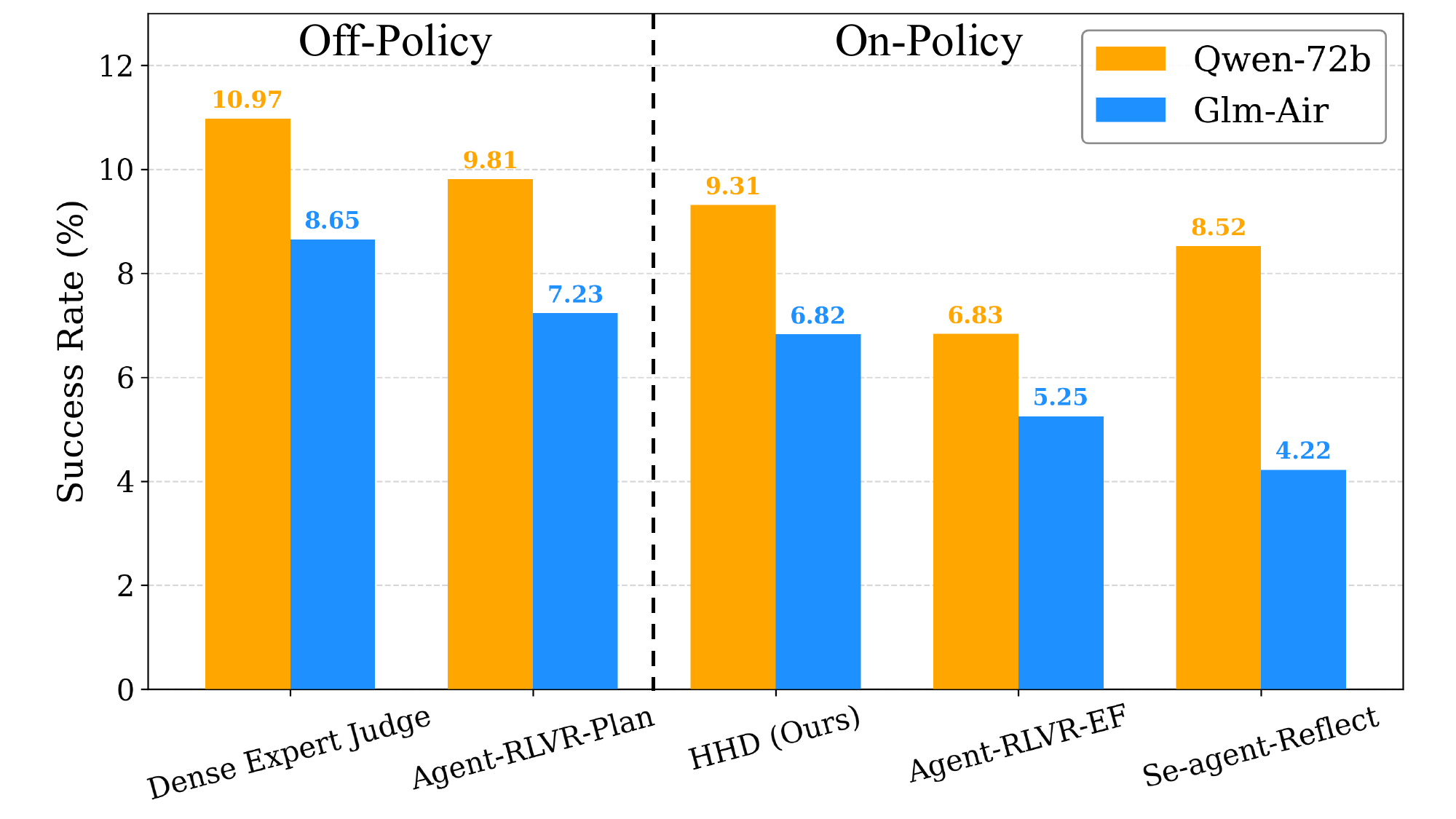} 
    \caption{Rollout efficiency of Qwen-72b and Glm-Air across different methods on previously failed tasks. The dashed line separates off-policy (left) and on-policy (right) algorithms. HHD achieves the highest  among the on-policy methods.}
    \label{fig:rollout_comparison}
    \vspace{-20pt}  
\end{figure}

\textbf{Rollout Efficiency Analysis.}
Besides trajectory quality, we examine the rollout success rates on previously failed tasks. As shown in Figure~\ref{fig:rollout_comparison}, HHD demonstrates robust rollout efficiency, achieving success rates comparable to the top-performing baselines. 

Crucially, we attribute the slightly higher success rates of baselines like \textit{Dense Expert Judge} and \textit{Agent-RLVR-Plan} to their reliance on \textbf{off-policy} mechanisms. While these methods may steer exploration towards successful outcomes, the resulting data is less effective for model training due to significant distribution shifts. 

\textbf{Dense Expert Judge (Step-Level Rejection Sampling):} The Dense Expert Judge creates an \textit{improved} distribution $\pi^{*}$ by applying an external expert as a judge to the base policy's proposals (rejection sampling). This process acts similarly to the selection phase in Monte Carlo Tree Search (MCTS), reshaping the sampling probability:
\begin{equation}
    \pi^{*}(a_t \mid s_t) \propto \pi_{\theta}(a_t \mid s_t) \cdot \exp\left(Q_{E}(s_t, a_t)\right)
\end{equation}
where $Q_{E}$ represents the preference of the expert judge. By suppressing suboptimal actions and promoting optimal ones, the Dense Expert Judge baseline forces the model $\pi$ to fit a highly divergent distribution $\pi^{*}$. This creates a severe off-policy distribution shift, making the target trajectories difficult to internalize.

\textbf{Agent-RLVR-Plan (Middle Injection):} Methods such as Agent-RLVR-Plan employ a mid-trajectory intervention strategy. Specifically, an expert equipped with privileged information (e.g., the ground truth patch) intervenes to ``take over'' the generation process, guiding the model to continue the rollout from a more optimal state. Training on such data creates distributional mismatch: the model is required to predict the expert's takeover conditioned on it's own past generation, which violates autoregressive consistency. This forces the model to memorize expert reasoning trace, thereby hindering its ability to generalize.


Unlike these approaches, HHD ensures the data distribution aligns perfectly with the model's policy. By conditioning on a global hint provided at the start, the entire trajectory is generated autoregressively by the model itself, making the reasoning patterns significantly easier to learn.

\textbf{Summary.} HHD outperforms baselines by generating high-quality, on-policy trajectories that significantly benefit learning. Notably, it maintains competitive rollout efficiency comparable to off-policy methods. This balance of trajectory quality, on-policyness, and efficiency leads to the significant performance gains observed in our experiments.

\subsection{Ablation Studies}
\label{sec:ablation}

In this section, we validate two core design choices of HHD: the position of hint injection and the necessity of explicit hint synthesis.

\begin{figure}[t]
    \centering
    \begin{minipage}[t]{0.48\columnwidth}
        \centering
        \includegraphics[width=\linewidth]{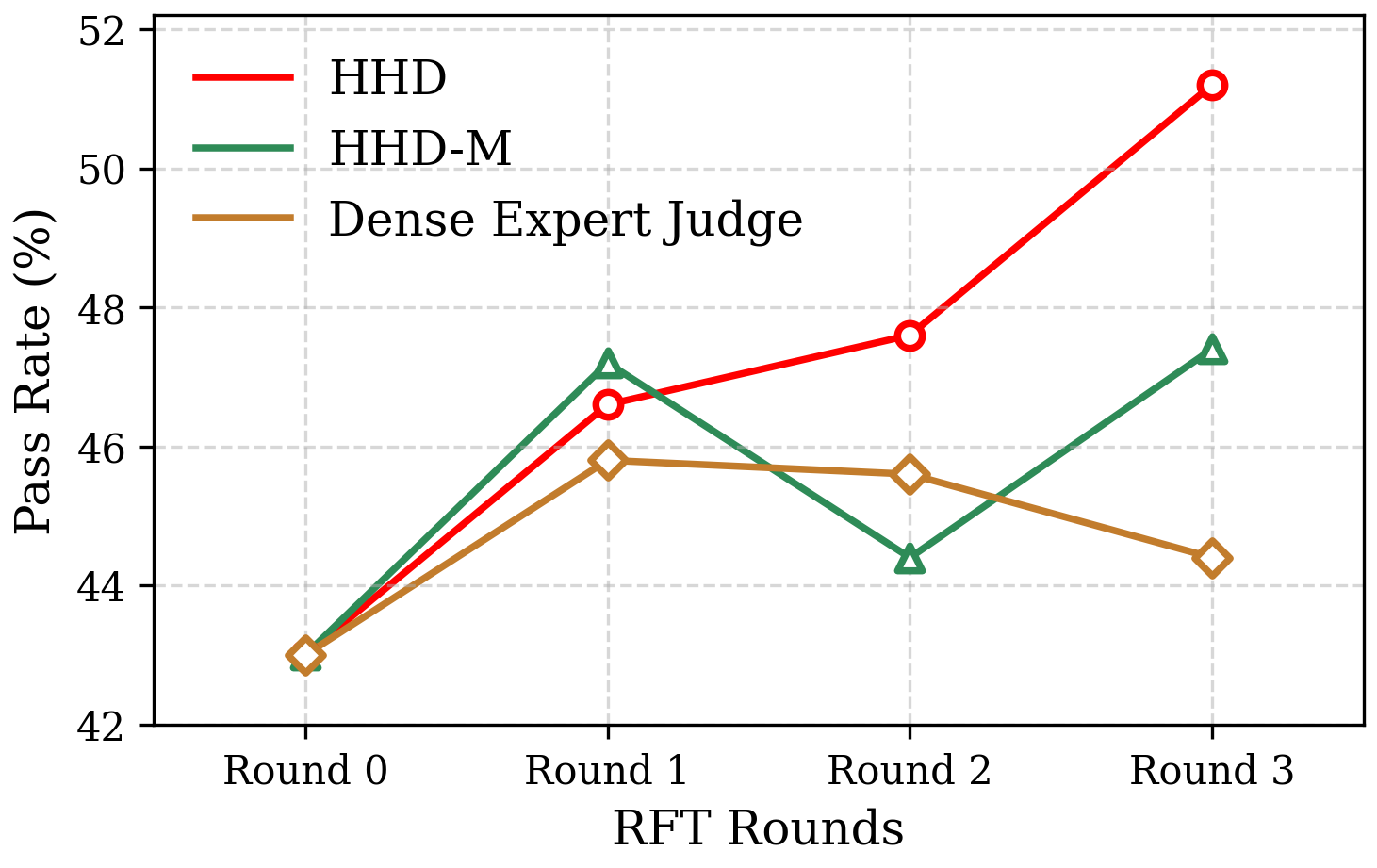}
        \caption*{(a) Qwen-72b ablation}
    \end{minipage}
    \hfill
    \begin{minipage}[t]{0.48\columnwidth}
        \centering
        \includegraphics[width=\linewidth]{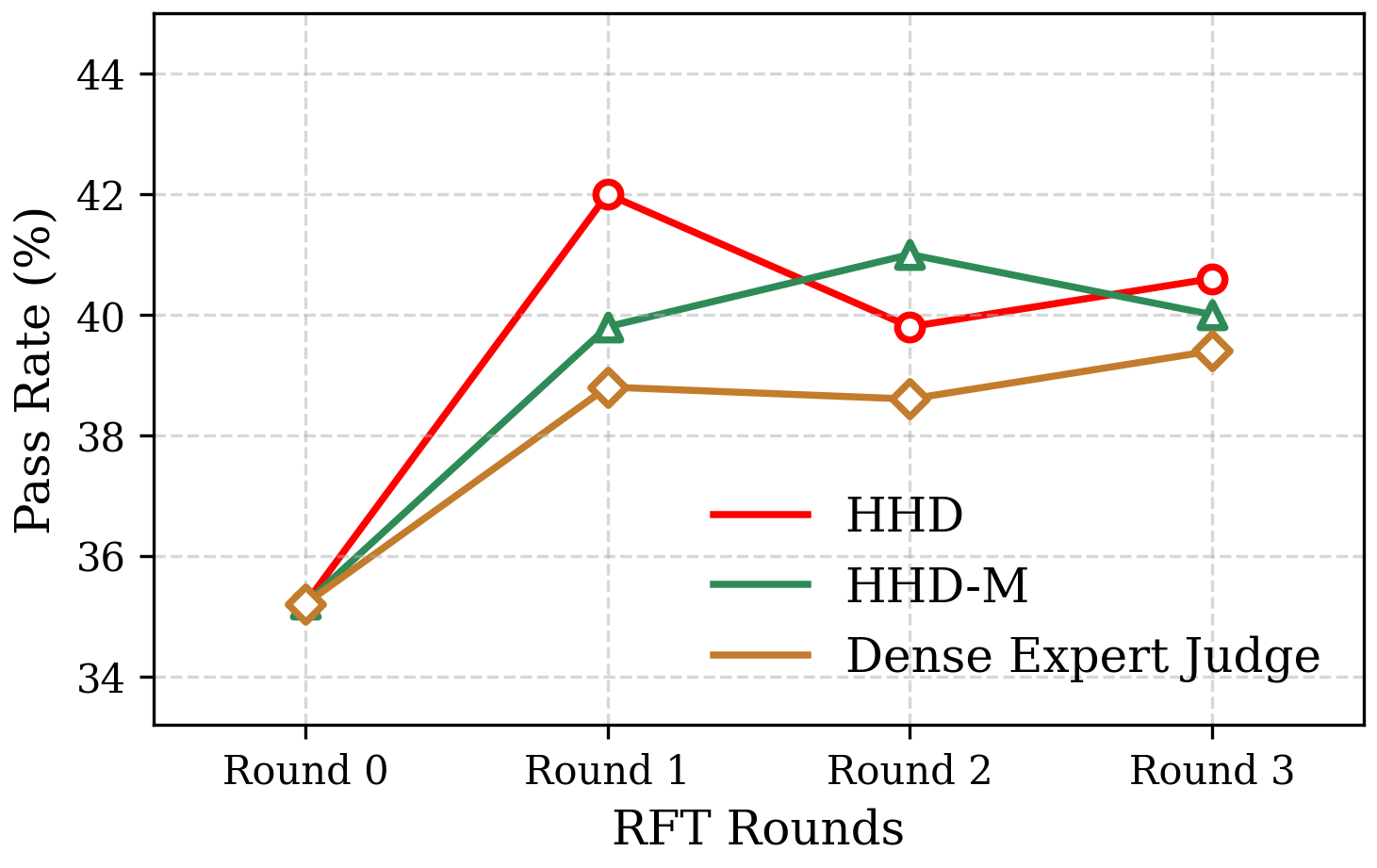}
        \caption*{(b) GLM-Air ablation}
    \end{minipage}
    \caption{Ablation study of HHD on Qwen-72b and GLM-Air models, showing that introducing the hint at the beginning of the trajectory yields the best performance.}
    \label{fig:ablation}
    \vspace{-15pt}
\end{figure}

\textbf{Impact of Hint Injection Position.} We compare HHD with a baseline variant, \textbf{HHD-M} (Middle). HHD-M adopts the mid-trajectory injection paradigm similar to the Agent-RLVR-Plan baseline. However, unlike the baseline, HHD-M injects the specific \textit{hindsight hint} synthesized by our method rather than external plans. As shown in Figure~\ref{fig:ablation}, HHD consistently outperforms HHD-M. We attribute this to the following: 1) \textbf{Distributional Mismatch:} As analyzed in Section~\ref{sec:analysis}, directly training on such data suffer from distributional mismatch, forcing the model to memorize expert reasoning trace, rather than learning generalizable logic. 2) \textbf{Missed Early Guidance:} Late injection fails to guide the initial steps. Consequently, the agent may waste effort on incorrect paths or commit to suboptimal strategies before the hint becomes available.

In contrast, HHD incorporates the hint \textbf{at the onset}, preventing early deviation and ensuring the policy is globally conditioned throughout the episode.

\textbf{Necessity of Explicit Hint Synthesis.} To assess the role of explicit hint generation and guidance, we compare HHD with the Dense Expert Judge. As shown in Section~\ref{sec:analysis}, while the latter achieves higher rollout efficiency through implicit step-level guidance, it introduces a critical limitation: the generated trajectories are \textit{off-policy}. By selecting optimal actions via rejection sampling at every reasoning step, the Dense Expert Judge creates a distribution mismatch between the training data and the model's natural behavior. HHD addresses this by using the explicit hint $h$ to globally condition the generation. This ensures the entire trajectory is produced autoregressively by the model itself, maintaining strict on-policyness and yielding higher-quality training signals that are easier for the model to internalize.

\section{Conclusion}

In this work, we introduce Hindsight Hint Distillation (HHD), a hint-based learning framework inspired by human teaching strategies, to address the challenge of improving reasoning on complex, long-horizon tasks. By enabling models to autonomously derive improved reasoning trajectories from large-scale CoT-free question--answer data, HHD mitigates the limitations of conventional fine-tuning and self-improvement methods that rely primarily on unguided exploration. 
Experiments on SWE-bench demonstrate that HHD achieves substantial performance gains over iterative RFT and existing trajectory synthesis baselines, while also exhibiting stronger generalization to multilingual settings. These results highlight the effectiveness of guided self-reflection through targeted hints in internalizing reasoning strategies, and suggest that HHD offers a scalable and practical approach for training models with enhanced problem-solving and long-horizon reasoning capabilities.

\section*{Acknowledgements}

This work was completed while the co-first authors were interns at Moonshot AI. 
The authors gratefully acknowledge Moonshot AI for providing the training infrastructure, computational resources, and research support that made this work possible. 
The public release of this paper has been reviewed and approved by Moonshot AI.

\nocite{langley00}

\bibliography{example_paper}
\bibliographystyle{icml2026}

\newpage
\appendix
\onecolumn
\section*{Appendix}
\section{Implementation of HHD}
\label{app:impl_hhd}
As described in the main text, HHD consists of four stages, among which hint generation from failed trajectories plays a central role. In this section, we detail the implementation of the hint generation procedure.

To reduce context length, we first compress each original trajectory into a compact format, as illustrated in Listing~\ref{lst:comp_traj}. For excessively long observations, most content is omitted, since in practice the quality of a trajectory can be effectively assessed by examining the sequence of actions produced by the model.

The compressed trajectory, together with the corresponding ground-truth patch, is then provided as input to an expert model (DeepSeek v3), which produces textual hints identifying the causes of failure. The full prompt used for hint generation is shown in Listing~\ref{lst:hdd_prompt}. Notably, the prompt is domain-agnostic, enabling applicability across a wide range of multi-turn reasoning tasks.

Finally, the generated hints are concatenated with the original problem statement, as illustrated in Listing~\ref{lst:hint_problem}, where the bold text corresponds to the inserted hints.

\begin{lstlisting}[caption={Example of one step in a compressed trajectory.}, label={lst:comp_traj}][t]

(*@\textcolor{black}{\textbf{Step 1}}@*)
"step":1
"action":"think(thought=Let me break down the issue:

1. The problem is about partial object instantiation in Hydra
2. Currently, if a config group has a missing value, partial instantiation fails with an exception
3. The request is to instead return a partial object where missing values become required arguments of the partial object
4. This is for use in hyperparameter search systems where parameters are proposed later

Key technical details:
- The code uses `@dataclass` with `MISSING` from `omegaconf`
- The config has `_target_` and `_partial_` fields
- The goal is to allow `hydra.utils.instantiate(cfg)` to work even with missing values

The example shows:
- A PartialFoo class with `foo: bool = MISSING`
- When `_partial_: bool = True`, it should return a partial object instead of throwing an exception
- The missing value should become a required argument in the partial object

This is a feature request, so I need to:
1. Understand the current behavior
2. Find where partial instantiation is handled
3. Modify the logic to handle missing values gracefully
4. Ensure the partial object can be completed later with the missing values)"
"observation":"EXECUTION RESULT of [think]:
Your thought has been logged."

(*@\textcolor{black}{\textbf{......}}@*)

(*@\textcolor{black}{\textbf{Step 9}}@*)

"step":9
"action":"str_replace_editor(command=view, path=/testbed/hydra/_internal/instantiate/_instantiate2.py, view_range=[270, 361])"
"observation":"EXECUTION RESULT of [str_replace_editor]:
Here's the result of running `cat -n` on /testbed/hydra/_i..."

(*@\textcolor{black}{\textbf{......}}@*)
\end{lstlisting}

\begin{lstlisting}[caption={Prompt template of hint generation. Prompt template for the expert model, where \texttt{\{problem\_statement\}} is the problem description, \texttt{\{gt\_patch\}} is the ground-truth patch, and \texttt{\{compressed\_trajectory\}} is the compressed action trajectory.}, label={lst:hdd_prompt}][t]
(*@\textcolor{black}{\textbf{System Prompt:}}@*)
"""You are an expert in analyzing agent trajectories.
Your task is to generate a **short hint** (2-3 sentences) that clarifies how the original problem statement should be improved, using insights from the original problem statement, the ground truth patch, and the trajectory.

Definitions:  
- **Original problem statement**: the initial description of what needs to be fixed or implemented.  
- **Ground truth patch**: the correct modification that fully solves the problem.  
- **Trajectory**: the sequence of steps the agent actually performed.  

Your job:  
1. Compare the original problem statement with the ground truth patch.  
2. Use the trajectory to identify where the original statement might have been unclear or misleading.  
3. Generate a **concise hint** that strengthens or clarifies the original problem statement.  

Requirements:  
- The hint must NOT explicitly mention the trajectory, ground truth patch, or agent actions. 
- The hint should be very short (2-3 sentences).
- Output must be JSON with a single key: `hint`.  

---
## Output Example
```json
{
  "hint": "Make sure the fix is applied specifically in function X. Clarify that the new parameter should have a default value."
}
```
"""

(*@\textcolor{black}{\textbf{User Prompt:}}@*)
"""
Problem statement: {problem_statement}
Ground truth patch: {gt_patch}
Trajectory: {compressed_trajectory}
"""
\end{lstlisting}

\begin{lstlisting}[
  caption={Example of a hint-augmented problem statement (``pandas-dev\_\_pandas-47643'' instance in SWE-Gym dataset).},
  label={lst:hint_problem},
  escapeinside={(*@}{@*)},
  basicstyle=\ttfamily\small,
]
<issue_description>
ENH: add suffixes argument to DataFrame.compare
#### Is your feature request related to a problem?

It would be convenient to choose other suffixes in `DataFrame.compare` than `self, other`

#### Describe the solution you'd like

Add argument `suffixes` so users can specify them.

#### API breaking implications

None

#### Additional context

```python
df = pd.DataFrame(
    {"col1": ["a", "b", "c"], "col2": [1.0, 2.0, np.nan], "col3": [1.0, 2.0, 3.0]},
    columns=["col1", "col2", "col3"],
)
df2 = df.copy()
df2.loc[0, "col1"] = "c"
df2.loc[2, "col3"] = np.nan
diff = df.compare(df2, suffixes=["left", "right"])
print(diff)

  col1       col3      
  left right left right
0    a     c  NaN   NaN
2  NaN   NaN  3.0   NaN
```
(*@\textcolor{black}{\textbf{Hint:}}@*)
Clarify that the new argument should be named 'result_names' instead of 'suffixes' and specify that it should accept a tuple of strings with default values ('self', 'other').

</issue_description>
\end{lstlisting}

\section{Details of Case Study}
\label{sec:appendix_detail_of_case_study}

A case study from SWE-bench Verified. Consistent with Section~\ref{sec:analysis}, the baseline exhibits \textit{suboptimal reasoning} and \textit{inefficient trial-and-error}. In contrast, HHD mitigates these issues through more effective high-level reasoning. This capability is acquired during training, where the model \textbf{internalizes} strategies from hint-guided rollouts and subsequently \textbf{generalizes} them to unseen tasks (e.g., this Django case) without explicit guidance at inference time.

\begin{table*}[h]
    \centering
    \small
    \renewcommand{\arraystretch}{1.5}
    \caption{ A case study from SWE-bench Verified. } 
    \label{tab:case_study_django}
    
    \begin{tabularx}{\textwidth}{@{} X|X @{}}
        \toprule
        \multicolumn{2}{c}{\textbf{Task: Enable Direct Iteration for Paginator (Django)}} \\
        \multicolumn{2}{p{0.96\textwidth}}{
        \textbf{Goal:} Implement \texttt{\_\_iter\_\_} to allow Pythonic looping (e.g., \texttt{for page in paginator}) without breaking backward compatibility. \newline
        } \\
        \midrule
        \multicolumn{1}{c|}{\textbf{\textcolor{red}{\ding{55} Failure Trajectory (Baseline)}}} & \multicolumn{1}{c}{\textbf{\textcolor{teal}{\ding{51} Success Trajectory (HHD)}}} \\
        \midrule
        
        \textbf{1. Suboptimal Reasoning: Reinventing the Wheel} \newline
        The agent failed to grasp the high-level intent. It attempted to \textbf{reinvent the wheel} by manually calculating indices and slicing the object list, ignoring the existing \texttt{page()} method.
        & 
        \textbf{1. Optimal Reasoning: Leveraging Abstractions} \newline
        The agent correctly identified that \texttt{\_\_iter\_\_} is just a wrapper. It formulated a plan to simply iterate over \texttt{self.page\_range} and yield \texttt{self.page(num)}.
        \\ \addlinespace[1em] 
        
        \textbf{2. Process Trap: Inefficient Debugging Loop} \newline
        The agent got locked into a \textbf{``Local Optima'' loop}. It spent numerous steps debugging low-level off-by-one errors and edge cases (e.g., empty lists) unrelated to the task, unable to break free and reassess the global context.
        & 
        \textbf{2. Process Pivot: Isolated Verification} \newline
        When faced with environment errors unrelated to the task, the agent distinguished noise from the core issue. It adhered to the high-level plan by constructing a targeted script, allowing it to focus exclusively on the relevant bug without getting trapped in unrelated system failures.
        \\ \addlinespace[1em]
        
        \textbf{3. Outcome: Incorrect and Bloated Code} \newline
        The trajectory was filled with inefficient exploration. The final code was bloated, fragile, and failed to handle \textbf{boundary conditions}, leading to regressions.
        & 
        \textbf{3. Outcome: Correct and Concise Solution} \newline
        The model solved the task with a concise \textbf{4-line fix}. It successfully internalized the strategy of ``reusing abstractions'' rather than memorizing a specific answer.
        \\        
        \bottomrule
        
    \end{tabularx}
\end{table*}


\subsection{Baseline Failed Trajectory}

We here provide the detailed trajectory for the failure case study referenced in Table \ref{tab:case_study_django}.

\label{sec:case_study_failed}

\paragraph{Phase 1: Suboptimal Reasoning.} 
The agent's initial strategy relies on heuristic assumptions rather than environmental verification. In Step 9, it attempts to use \texttt{pytest}—a standard tool in modern Python but not configured in this specific environment—without first checking the \texttt{README} or available test runners. 

\begin{lstlisting}[caption={Suboptimal Reasoning. The agent blindly assumes tool availability.}, label={lst:misalignment}, escapeinside={(*@}{@*)}]
(*@\textcolor{black}{\textbf{ACTION (Step 9):}}@*)
<function=execute_bash>
<parameter=command>cd /testbed && python -m pytest tests/pagination/tests.py::PaginationTests::test_page_range_iterator -v</parameter>
</function>
(*@\textcolor{black}{\textbf{OBSERVATION:}}@*)
EXECUTION RESULT of [execute_bash]:
/opt/miniconda3/envs/testbed/bin/python: No module named pytest
\end{lstlisting}

\paragraph{Phase 2: The Local Optima Trap.} 
Instead of pivoting to a high-level planning state to investigate the correct test runner (e.g., \texttt{runtests.py} or the specific Django test command structure), the agent devolves into an inefficient trial-and-error loop. As shown in Listing~\ref{lst:trap}, the agent repeatedly permutes command arguments (Steps 10--13), eventually leading to termination by the environment due to the detected loop.


\begin{lstlisting}[caption={The Local Optima Trap. The agent iterates on a flawed approach (Inefficient Trial-and-Error) rather than diagnosing the root cause.}, label={lst:trap}, escapeinside={(*@}{@*)}]
(*@\textcolor{black}{\textbf{ACTION (Step 10):}}@*)
<function=execute_bash>
<parameter=command>cd /testbed && python -m django.test.runner discover tests/pagination/tests.py::PaginationTests::test_page_range_iterator</parameter>
</function>
(*@\textcolor{black}{\textbf{OBSERVATION:}}@*)
Your command ran successfully and did not produce any output.

(*@\textcolor{black}{\textbf{ACTION (Step 11):}}@*)
<function=execute_bash>
<parameter=command>cd /testbed && python -m django.test.runner discover tests/pagination/tests.py -v</parameter>
</function>
(*@\textcolor{black}{\textbf{OBSERVATION:}}@*)
Your command ran successfully and did not produce any output.

(*@\textcolor{black}{\textbf{ACTION (Step 12):}}@*)
(*@\textit{...[Agent repeats similar command variant]}...@*)
<parameter=command>cd /testbed && python -m django.test.runner discover tests/pagination/tests.py::PaginationTests -v</parameter>
(*@\textcolor{black}{\textbf{OBSERVATION:}}@*)
Your command ran successfully and did not produce any output.
\end{lstlisting}

\subsection{HHD Success Trajectory}

We here provide the detailed trajectory for the success case study referenced in Table \ref{tab:case_study_django}.

\paragraph{Phase 1: Exploration and Localization.} 
The agent first analyzes the request to implement \texttt{\_\_iter\_\_} for the \texttt{Paginator} class. It locates the relevant file to understand the current implementation.

\begin{lstlisting}[caption={Exploration and file analysis.}, label={lst:explore}, escapeinside={(*@}{@*)}]
(*@\textcolor{black}{\textbf{ACTION:}}@*)
<function=think>
...I need to implement __iter__ to make it more Pythonic...
</function>
(*@\textcolor{black}{\textbf{OBSERVATION:}}@*)
Your thought has been logged.

(*@\textcolor{black}{\textbf{ACTION:}}@*)
<function=str_replace_editor>
<parameter=command>view</parameter>
<parameter=path>/testbed/django/core/paginator.py</parameter>
</function>
(*@\textcolor{black}{\textbf{OBSERVATION:}}@*)
...
class Paginator:
    ...
    @property
    def page_range(self):
        return range(1, self.num_pages + 1)
...
\end{lstlisting}

\paragraph{Phase 2: Reproduction.} 
The agent creates a test script to reproduce the current behavior.

\begin{lstlisting}[caption={Creating and running a reproduction script.}, label={lst:repro}, escapeinside={(*@}{@*)}]
(*@\textcolor{black}{\textbf{ACTION:}}@*)
<function=str_replace_editor>
<parameter=command>create</parameter>
<parameter=path>/testbed/test_iteration.py</parameter>
<parameter=file_text>
...
def test_current_behavior():
    """Test how Paginator currently works"""
    print("=== Testing Current Behavior ===")
...
</parameter>
</function>
(*@\textcolor{black}{\textbf{OBSERVATION:}}@*)
File created successfully at: /testbed/test_iteration.py

(*@\textcolor{black}{\textbf{ACTION:}}@*)
<function=execute_bash>
<parameter=command>cd /testbed && python test_iteration.py</parameter>
</function>
(*@\textcolor{black}{\textbf{OBSERVATION:}}@*)
=== Testing Current Behavior ===
Total items: 20
Items per page:...
\end{lstlisting}

\paragraph{Phase 3: Implementation.} 
The agent modifies the \texttt{Paginator} class by adding the \texttt{\_\_iter\_\_} method.

\begin{lstlisting}[caption={Implementing the \texttt{\_\_iter\_\_} method.}, label={lst:implementation}, escapeinside={(*@}{@*)}]
(*@\textcolor{black}{\textbf{ACTION:}}@*)
<function=str_replace_editor>
<parameter=command>str_replace</parameter>
<parameter=path>/testbed/django/core/paginator.py</parameter>
<parameter=old_str>
    @property
    def page_range(self):
</parameter>
<parameter=new_str>
    def __iter__(self):
        for page_num in self.page_range:
            yield self.page(page_num)

    @property
    def page_range(self):
</parameter>
</function>
(*@\textcolor{black}{\textbf{OBSERVATION:}}@*)
The file /testbed/django/core/paginator.py has been edited...
\end{lstlisting}

\paragraph{Phase 4: Verification.} 
Finally, the agent verifies the fix using a script that iterates through the paginator and prints the objects.

\begin{lstlisting}[caption={Verifying the fix.}, label={lst:verify}, escapeinside={(*@}{@*)}]
(*@\textcolor{black}{\textbf{ACTION:}}@*)
<function=execute_bash>
<parameter=command>cd /testbed && python -c "
...
print('New way:')
for page in paginator:
    print('  Page {}: {}'.format(page.number, list(page.object_list)))
"
</parameter>
</function>
(*@\textcolor{black}{\textbf{OBSERVATION:}}@*)
New way:
  Page 1: [1, 2, 3, 4, 5]
  Page 2: [6, 7, 8, 9, 10]
  ...
\end{lstlisting}

\section{Pipeline of Failure Case Study}

\begin{figure}[htbp]
    \centering
    \includegraphics[width=\linewidth]{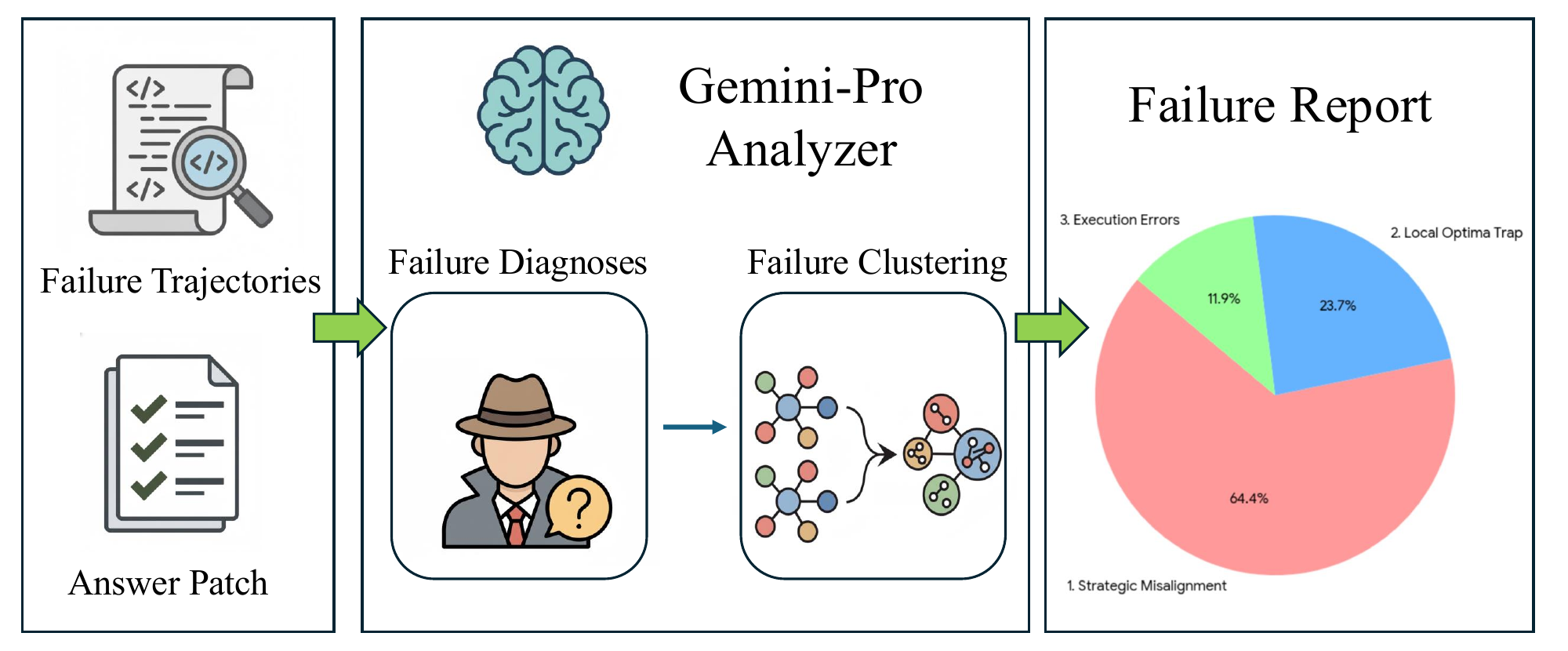} 
    \caption{\textbf{Overview of the automated failure diagnosis pipeline.} The pipeline employs a frontier model (Gemini-Pro) to conduct root-cause analysis by examining agent interaction histories, ground-truth patches, and execution feedback, and categorizes systematic failure modes.}
    \label{fig:failure_pipeline}
\end{figure}

To systematically analyze the failure patterns of the base model on SWE-bench, we construct an \textbf{automated failure diagnosis pipeline} for root-cause analysis of agent failures. As shown in Figure~\ref{fig:failure_pipeline}, we collect 60 failure trajectories generated by the baseline model on SWE-bench. For each instance, a frontier model (Gemini-Pro) is employed as an impartial evaluator. The evaluator examines the agent’s full interaction history, the corresponding ground-truth patch, and test execution feedback to identify where and why the agent deviates from the correct solution. 
Based on this analysis, we cluster the identified failure modes and categorize them into three primary types: \textit{strategic misalignment}, \textit{stagnation in local optima}, and \textit{execution errors}.

\section{Details of Experiments}
\label{app:exp}

\subsection{Agentic Training Details}
\label{app:models}
To activate tool usage within the OpenHands framework, we bootstrap agentic behavior via trajectory-based supervision. We use the Kimi-K2 model~\cite{team2025kimi} to perform tool-augmented rollouts on the SWE-Gym dataset, executing two full passes over the problem set. Each rollout produces complete interaction trajectories involving OpenHands actions.
Reject-sampling is applied to the collected trajectories, retaining only successful trajectories for fine-tuning the base model. Training is performed for 3 epochs with a global batch size of 64 and a maximum sequence length of 128K tokens, supporting long-context reasoning for software engineering tasks. We adopt a cosine learning rate schedule with an initial learning rate of $1\times10^{-5}$ and a minimum learning rate of $1\times10^{-6}$.
This procedure enables the model to internalize effective tool-usage patterns without explicit tool-specific supervision.

\subsection{Details of the Agentic Framework}
\label{app:openhands}

This section describes the OpenHands scaffold, including its action space and the concrete system and user prompts used by the agent.

\subsubsection{OpenHands Action Space}

The system prompt defines a set of JSON-style function-calling actions:
\begin{itemize}[noitemsep, topsep=0pt]
    \item \texttt{bash}: Executes shell commands in the environment.
    \item \texttt{think}: Records intermediate reasoning or planning steps without producing environmental side effects.
    \item \texttt{str\_replace\_editor}: Views file contents and performs targeted code edits.
    \item \texttt{finish}: Produces a final response and marks the task as complete.
\end{itemize}
Observations capture environmental feedback (e.g., execution logs and file updates) and are used to inform subsequent agent actions.

\subsection{OpenHands System and User Prompts}

The system and user prompts are provided in Listings~\ref{lst:sys_prompt} and~\ref{lst:user_prompt}, respectively. For each problem instance, the problem description is injected into the user prompt by replacing the placeholder enclosed between $\langle$issue\_description$\rangle$ and $\langle$/issue\_description$\rangle$.

\begin{lstlisting}[caption={System prompt of OpenHands scaffold.}, label={lst:sys_prompt}][t]
You are OpenHands agent, a helpful AI assistant that can interact with a computer to solve tasks.

<ROLE>
Your primary role is to assist users by executing commands, modifying code, and solving technical problems effectively. You should be thorough, methodical, and prioritize quality over speed.
* If the user asks a question, like "why is X happening", don't try to fix the problem. Just give an answer to the question.
</ROLE>

<EFFICIENCY>
* Each action you take is somewhat expensive. Wherever possible, combine multiple actions into a single action, e.g. combine multiple bash commands into one, using sed and grep to edit/view multiple files at once.
* When exploring the codebase, use efficient tools like find, grep, and git commands with appropriate filters to minimize unnecessary operations.
</EFFICIENCY>

<FILE_SYSTEM_GUIDELINES>
* When a user provides a file path, do NOT assume it's relative to the current working directory. First explore the file system to locate the file before working on it.
* If asked to edit a file, edit the file directly, rather than creating a new file with a different filename.
* For global search-and-replace operations, consider using `sed` instead of opening file editors multiple times.
</FILE_SYSTEM_GUIDELINES>

<CODE_QUALITY>
* Write clean, efficient code with minimal comments. Avoid redundancy in comments: Do not repeat information that can be easily inferred from the code itself.
* When implementing solutions, focus on making the minimal changes needed to solve the problem.
* Before implementing any changes, first thoroughly understand the codebase through exploration.
* If you are adding a lot of code to a function or file, consider splitting the function or file into smaller pieces when appropriate.
</CODE_QUALITY>

<VERSION_CONTROL>
* When configuring git credentials, use "openhands" as the user.name and "openhands@all-hands.dev" as the user.email by default, unless explicitly instructed otherwise.
* Exercise caution with git operations. Do NOT make potentially dangerous changes (e.g., pushing to main, deleting repositories) unless explicitly asked to do so.
* When committing changes, use `git status` to see all modified files, and stage all files necessary for the commit. Use `git commit -a` whenever possible.
* Do NOT commit files that typically shouldn't go into version control (e.g., node_modules/, .env files, build directories, cache files, large binaries) unless explicitly instructed by the user.
* If unsure about committing certain files, check for the presence of .gitignore files or ask the user for clarification.
</VERSION_CONTROL>

<PULL_REQUESTS>
* When creating pull requests, create only ONE per session/issue unless explicitly instructed otherwise.
* When working with an existing PR, update it with new commits rather than creating additional PRs for the same issue.
* When updating a PR, preserve the original PR title and purpose, updating description only when necessary.
</PULL_REQUESTS>

<PROBLEM_SOLVING_WORKFLOW>
1. EXPLORATION: Thoroughly explore relevant files and understand the context before proposing solutions
2. ANALYSIS: Consider multiple approaches and select the most promising one
3. TESTING:
   * For bug fixes: Create tests to verify issues before implementing fixes
   * For new features: Consider test-driven development when appropriate
   * If the repository lacks testing infrastructure and implementing tests would require extensive setup, consult with the user before investing time in building testing infrastructure
   * If the environment is not set up to run tests, consult with the user first before investing time to install all dependencies
4. IMPLEMENTATION: Make focused, minimal changes to address the problem
5. VERIFICATION: If the environment is set up to run tests, test your implementation thoroughly, including edge cases. If the environment is not set up to run tests, consult with the user first before investing time to run tests.
</PROBLEM_SOLVING_WORKFLOW>

<SECURITY>
* Only use GITHUB_TOKEN and other credentials in ways the user has explicitly requested and would expect.
* Use APIs to work with GitHub or other platforms, unless the user asks otherwise or your task requires browsing.
</SECURITY>

<ENVIRONMENT_SETUP>
* When user asks you to run an application, don't stop if the application is not installed. Instead, please install the application and run the command again.
* If you encounter missing dependencies:
  1. First, look around in the repository for existing dependency files (requirements.txt, pyproject.toml, package.json, Gemfile, etc.)
  2. If dependency files exist, use them to install all dependencies at once (e.g., `pip install -r requirements.txt`, `npm install`, etc.)
  3. Only install individual packages directly if no dependency files are found or if only specific packages are needed
* Similarly, if you encounter missing dependencies for essential tools requested by the user, install them when possible.
</ENVIRONMENT_SETUP>

<TROUBLESHOOTING>
* If you've made repeated attempts to solve a problem but tests still fail or the user reports it's still broken:
  1. Step back and reflect on 5-7 different possible sources of the problem
  2. Assess the likelihood of each possible cause
  3. Methodically address the most likely causes, starting with the highest probability
  4. Document your reasoning process
* When you run into any major issue while executing a plan from the user, please don't try to directly work around it. Instead, propose a new plan and confirm with the user before proceeding.
</TROUBLESHOOTING>
You have access to the following functions:

---- BEGIN FUNCTION #1: execute_bash ----
Description: Execute a bash command in the terminal within a persistent shell session.


### Command Execution
* One command at a time: You can only execute one bash command at a time. If you need to run multiple commands sequentially, use `&&` or `;` to chain them together.
* Persistent session: Commands execute in a persistent shell session where environment variables, virtual environments, and working directory persist between commands.
* Soft timeout: Commands have a soft timeout of 10 seconds, once that's reached, you have the option to continue or interrupt the command (see section below for details)

### Long-running Commands
* For commands that may run indefinitely, run them in the background and redirect output to a file, e.g. `python3 app.py > server.log 2>&1 &`.
* For commands that may run for a long time (e.g. installation or testing commands), or commands that run for a fixed amount of time (e.g. sleep), you should set the "timeout" parameter of your function call to an appropriate value.
* If a bash command returns exit code `-1`, this means the process hit the soft timeout and is not yet finished. By setting `is_input` to `true`, you can:
  - Send empty `command` to retrieve additional logs
  - Send text (set `command` to the text) to STDIN of the running process
  - Send control commands like `C-c` (Ctrl+C), `C-d` (Ctrl+D), or `C-z` (Ctrl+Z) to interrupt the process
  - If you do C-c, you can re-start the process with a longer "timeout" parameter to let it run to completion

### Best Practices
* Directory verification: Before creating new directories or files, first verify the parent directory exists and is the correct location.
* Directory management: Try to maintain working directory by using absolute paths and avoiding excessive use of `cd`.

### Output Handling
* Output truncation: If the output exceeds a maximum length, it will be truncated before being returned.

Parameters:
  (1) command (string, required): The bash command to execute. Can be empty string to view additional logs when previous exit code is `-1`. Can be `C-c` (Ctrl+C) to interrupt the currently running process. Note: You can only execute one bash command at a time. If you need to run multiple commands sequentially, you can use `&&` or `;` to chain them together.
  (2) is_input (string, optional): If True, the command is an input to the running process. If False, the command is a bash command to be executed in the terminal. Default is False.
Allowed values: [`true`, `false`]
  (3) timeout (number, optional): Optional. Sets a hard timeout in seconds for the command execution. If not provided, the command will use the default soft timeout behavior.
---- END FUNCTION #1 ----

---- BEGIN FUNCTION #2: think ----
Description: Use the tool to think about something. It will not obtain new information or make any changes to the repository, but just log the thought. Use it when complex reasoning or brainstorming is needed.

Common use cases:
1. When exploring a repository and discovering the source of a bug, call this tool to brainstorm several unique ways of fixing the bug, and assess which change(s) are likely to be simplest and most effective.
2. After receiving test results, use this tool to brainstorm ways to fix failing tests.
3. When planning a complex refactoring, use this tool to outline different approaches and their tradeoffs.
4. When designing a new feature, use this tool to think through architecture decisions and implementation details.
5. When debugging a complex issue, use this tool to organize your thoughts and hypotheses.

The tool simply logs your thought process for better transparency and does not execute any code or make changes.
Parameters:
  (1) thought (string, required): The thought to log.
---- END FUNCTION #2 ----

---- BEGIN FUNCTION #3: finish ----
Description: Signals the completion of the current task or conversation.

Use this tool when:
- You have successfully completed the user's requested task
- You cannot proceed further due to technical limitations or missing information

The message should include:
- A clear summary of actions taken and their results
- Any next steps for the user
- Explanation if you're unable to complete the task
- Any follow-up questions if more information is needed

The task_completed field should be set to True if you believed you have completed the task, and False otherwise.

Parameters:
  (1) message (string, required): Final message to send to the user
  (2) task_completed (string, required): Whether you have completed the task.
Allowed values: [`true`, `false`, `partial`]
---- END FUNCTION #3 ----

---- BEGIN FUNCTION #4: str_replace_editor ----
Description: Custom editing tool for viewing, creating and editing files in plain-text format
* State is persistent across command calls and discussions with the user
* If `path` is a text file, `view` displays the result of applying `cat -n`. If `path` is a directory, `view` lists non-hidden files and directories up to 2 levels deep
* The following binary file extensions can be viewed in Markdown format: [".xlsx", ".pptx", ".wav", ".mp3", ".m4a", ".flac", ".pdf", ".docx"]. IT DOES NOT HANDLE IMAGES.
* The `create` command cannot be used if the specified `path` already exists as a file
* If a `command` generates a long output, it will be truncated and marked with `<response clipped>`
* The `undo_edit` command will revert the last edit made to the file at `path`
* This tool can be used for creating and editing files in plain-text format.


Before using this tool:
1. Use the view tool to understand the file's contents and context
2. Verify the directory path is correct (only applicable when creating new files):
   - Use the view tool to verify the parent directory exists and is the correct location

When making edits:
   - Ensure the edit results in idiomatic, correct code
   - Do not leave the code in a broken state
   - Always use absolute file paths (starting with /)

CRITICAL REQUIREMENTS FOR USING THIS TOOL:

1. EXACT MATCHING: The `old_str` parameter must match EXACTLY one or more consecutive lines from the file, including all whitespace and indentation. The tool will fail if `old_str` matches multiple locations or doesn't match exactly with the file content.

2. UNIQUENESS: The `old_str` must uniquely identify a single instance in the file:
   - Include sufficient context before and after the change point (3-5 lines recommended)
   - If not unique, the replacement will not be performed

3. REPLACEMENT: The `new_str` parameter should contain the edited lines that replace the `old_str`. Both strings must be different.

Remember: when making multiple file edits in a row to the same file, you should prefer to send all edits in a single message with multiple calls to this tool, rather than multiple messages with a single call each.

Parameters:
  (1) command (string, required): The commands to run. Allowed options are: `view`, `create`, `str_replace`, `insert`, `undo_edit`.
Allowed values: [`view`, `create`, `str_replace`, `insert`, `undo_edit`]
  (2) path (string, required): Absolute path to file or directory, e.g. `/workspace/file.py` or `/workspace`.
  (3) file_text (string, optional): Required parameter of `create` command, with the content of the file to be created.
  (4) old_str (string, optional): Required parameter of `str_replace` command containing the string in `path` to replace.
  (5) new_str (string, optional): Optional parameter of `str_replace` command containing the new string (if not given, no string will be added). Required parameter of `insert` command containing the string to insert.
  (6) insert_line (integer, optional): Required parameter of `insert` command. The `new_str` will be inserted AFTER the line `insert_line` of `path`.
  (7) view_range (array, optional): Optional parameter of `view` command when `path` points to a file. If none is given, the full file is shown. If provided, the file will be shown in the indicated line number range, e.g. [11, 12] will show lines 11 and 12. Indexing at 1 to start. Setting `[start_line, -1]` shows all lines from `start_line` to the end of the file.
---- END FUNCTION #4 ----


If you choose to call a function ONLY reply in the following format with NO suffix:

<function=example_function_name>
<parameter=example_parameter_1>value_1</parameter>
<parameter=example_parameter_2>
This is the value for the second parameter
that can span
multiple lines
</parameter>
</function>

<IMPORTANT>
Reminder:
- Function calls MUST follow the specified format, start with <function= and end with </function>
- Required parameters MUST be specified
- Only call one function at a time
- You may provide optional reasoning for your function call in natural language BEFORE the function call, but NOT after.
- If there is no function call available, answer the question like normal with your current knowledge and do not tell the user about function calls
</IMPORTANT>
\end{lstlisting}

\begin{lstlisting}[caption={User prompt of OpenHands scaffold.}, label={lst:user_prompt}][t]
Here's a running example of how to perform a task with the provided tools.

--------------------- START OF EXAMPLE ---------------------

USER: Create a list of numbers from 1 to 10, and display them in a web page at port 5000.


ASSISTANT: Sure! Let me first check the current directory:
<function=execute_bash>
<parameter=command>
pwd && ls
</parameter>
</function>

USER: EXECUTION RESULT of [execute_bash]:
/workspace
openhands@runtime:~/workspace$

ASSISTANT:
There is no `app.py` file in the current directory. Let me create a Python file `app.py`:
<function=str_replace_editor>
<parameter=command>create</parameter>
<parameter=path>/workspace/app.py</parameter>
<parameter=file_text>
from flask import Flask
app = Flask(__name__)

@app.route('/')
def index():
    numbers = list(range(1, 11))
    return str(numbers)

if __name__ == '__main__':
    app.run(port=5000)
</parameter>
</function>

USER: EXECUTION RESULT of [str_replace_editor]:
File created successfully at: /workspace/app.py

ASSISTANT:
Let me run the Python file for you:
<function=execute_bash>
<parameter=command>
python3 app.py > server.log 2>&1 &
</parameter>
</function>

USER: EXECUTION RESULT of [execute_bash]:
[1] 121
[1]+  Exit 1                  python3 app.py > server.log 2>&1

ASSISTANT:
Looks like the server was running with PID 121 then crashed. Let me check the server log:
<function=execute_bash>
<parameter=command>
cat server.log
</parameter>
</function>

USER: EXECUTION RESULT of [execute_bash]:
Traceback (most recent call last):
  File "/workspace/app.py", line 2, in <module>
    from flask import Flask
ModuleNotFoundError: No module named 'flask'

ASSISTANT:
Looks like the server crashed because the `flask` module is not installed. Let me install the `flask` module for you:
<function=execute_bash>
<parameter=command>
pip3 install flask
</parameter>
</function>

USER: EXECUTION RESULT of [execute_bash]:
Defaulting to user installation because normal site-packages is not writeable
Collecting flask
  Using cached flask-3.0.3-py3-none-any.whl (101 kB)
Collecting blinker>=1.6.2
  Using cached blinker-1.7.0-py3-none-any.whl (13 kB)
Collecting Werkzeug>=3.0.0
  Using cached werkzeug-3.0.2-py3-none-any.whl (226 kB)
Collecting click>=8.1.3
  Using cached click-8.1.7-py3-none-any.whl (97 kB)
Collecting itsdangerous>=2.1.2
  Using cached itsdangerous-2.2.0-py3-none-any.whl (16 kB)
Requirement already satisfied: Jinja2>=3.1.2 in /home/openhands/.local/lib/python3.10/site-packages (from flask) (3.1.3)
Requirement already satisfied: MarkupSafe>=2.0 in /home/openhands/.local/lib/python3.10/site-packages (from Jinja2>=3.1.2->flask) (2.1.5)
Installing collected packages: Werkzeug, itsdangerous, click, blinker, flask
Successfully installed Werkzeug-3.0.2 blinker-1.7.0 click-8.1.7 flask-3.0.3 itsdangerous-2.2.0

ASSISTANT:
The `flask` module is installed successfully. Let me run the Python file again:
<function=execute_bash>
<parameter=command>
python3 app.py > server.log 2>&1 &
</parameter>
</function>

USER: EXECUTION RESULT of [execute_bash]:
[1] 124

ASSISTANT:
Let me check the server log again:
<function=execute_bash>
<parameter=command>
cat server.log
</parameter>
</function>

USER: EXECUTION RESULT of [execute_bash]:
* Serving Flask app 'app'
 * Debug mode: off
WARNING: This is a development server. Do not use it in a production deployment. Use a production WSGI server instead.
 * Running on http://127.0.0.1:5000
Press CTRL+C to quit

ASSISTANT:
The server is running on port 5000 with PID 124. You can access the list of numbers by visiting http://127.0.0.1:5000. If you have any further questions, feel free to ask!

USER: Now kill the server, make it display the numbers in a table format.

ASSISTANT:
Sure! Let me stop the server first:
<function=execute_bash>
<parameter=command>
kill 124
</parameter>
</function>

USER: EXECUTION RESULT of [execute_bash]:
[1]+  Terminated              python3 app.py > server.log 2>&1

ASSISTANT:
Now let me display the numbers in a table format:
<function=str_replace_editor>
<parameter=command>str_replace</parameter>
<parameter=path>/workspace/app.py</parameter>
<parameter=old_str>return str(numbers)</parameter>
<parameter=new_str>return '<table>' + ''.join([f'<tr><td>{i}</td></tr>' for i in numbers]) + '</table>'</parameter>
</function>

USER: EXECUTION RESULT of [str_replace_editor]:
The file /workspace/app.py has been edited. Here's the result of running `cat -n` on a snippet of /workspace/app.py:
     3
     4  @app.route('/')
     5  def index():
     6      numbers = list(range(1, 11))
     7      return '<table>' + ''.join([f'<tr><td>{i}</td></tr>' for i in numbers]) + '</table>'
     8
     9  if __name__ == '__main__':
    10      app.run(port=5000)
Review the changes and make sure they are as expected. Edit the file again if necessary.


ASSISTANT:
Running the updated file:
<function=execute_bash>
<parameter=command>
python3 app.py > server.log 2>&1 &
</parameter>
</function>

USER: EXECUTION RESULT of [execute_bash]:
[1] 126

ASSISTANT:
The server is running on port 5000 with PID 126. You can access the list of numbers in a table format by visiting http://127.0.0.1:5000.

ASSISTANT:
The server is running on port 5000 with PID 126. You can access the list of numbers in a table format by visiting http://127.0.0.1:5000. Let me know if you have any further requests!
<function=finish>
<parameter=message>The task has been completed. The web server is running and displaying numbers 1-10 in a table format at http://127.0.0.1:5000.</parameter>
<parameter=task_completed>true</parameter>
</function>

--------------------- END OF EXAMPLE ---------------------

Do NOT assume the environment is the same as in the example above.

--------------------- NEW TASK DESCRIPTION ---------------------
<uploaded_files>
/testbed
</uploaded_files>

I've uploaded a python code repository in the directory testbed. Consider the following issue description:

<issue_description>
(*@\textcolor{black}{\textbf{Specific Problem Statement}}@*)
</issue_description>

Can you help me implement the necessary changes to the repository so that the requirements specified in the <issue_description> are met?
I've already taken care of all changes to any of the test files described in the <issue_description>. This means you DON'T have to modify the testing logic or any of the tests in any way!
Also the development Python environment is already set up for you (i.e., all dependencies already installed), so you don't need to install other packages.
Your task is to make the minimal changes to non-test files in the /testbed directory to ensure the <issue_description> is satisfied.

Follow these phases to resolve the issue:

Phase 1. READING: read the problem and reword it in clearer terms
   1.1 If there are code or config snippets. Express in words any best practices or conventions in them.
   1.2 Hightlight message errors, method names, variables, file names, stack traces, and technical details.
   1.3 Explain the problem in clear terms.
   1.4 Enumerate the steps to reproduce the problem.
   1.5 Hightlight any best practices to take into account when testing and fixing the issue

Phase 2. RUNNING: install and run the tests on the repository
   2.1 Follow the readme
   2.2 Install the environment and anything needed
   2.2 Iterate and figure out how to run the tests

Phase 3. EXPLORATION: find the files that are related to the problem and possible solutions
   3.1 Use `grep` to search for relevant methods, classes, keywords and error messages.
   3.2 Identify all files related to the problem statement.
   3.3 Propose the methods and files to fix the issue and explain why.
   3.4 From the possible file locations, select the most likely location to fix the issue.

Phase 4. TEST CREATION: before implementing any fix, create a script to reproduce and verify the issue.
   4.1 Look at existing test files in the repository to understand the test format/structure.
   4.2 Create a minimal reproduction script that reproduces the located issue.
   4.3 Run the reproduction script to confirm you are reproducing the issue.
   4.4 Adjust the reproduction script as necessary.

Phase 5. FIX ANALYSIS: state clearly the problem and how to fix it
   5.1 State clearly what the problem is.
   5.2 State clearly where the problem is located.
   5.3 State clearly how the test reproduces the issue.
   5.4 State clearly the best practices to take into account in the fix.
   5.5 State clearly how to fix the problem.

Phase 6. FIX IMPLEMENTATION: Edit the source code to implement your chosen solution.
   6.1 Make minimal, focused changes to fix the issue.

Phase 7. VERIFICATION: Test your implementation thoroughly.
   7.1 Run your reproduction script to verify the fix works.
   7.2 Add edge cases to your test script to ensure comprehensive coverage.
   7.3 Run existing tests related to the modified code to ensure you haven't broken anything.

8. FINAL REVIEW: Carefully re-read the problem description and compare your changes with the base commit.
   8.1 Ensure you've fully addressed all requirements.
   8.2 Run any tests in the repository related to:
     8.2.1 The issue you are fixing
     8.2.2 The files you modified
     8.2.3 The functions you changed
   8.3 If any tests fail, revise your implementation until all tests pass

Be thorough in your exploration, testing, and reasoning. It's fine if your thinking process is lengthy - quality and completeness are more important than brevity.
--------------------- END OF NEW TASK DESCRIPTION ---------------------

PLEASE follow the format strictly! PLEASE EMIT ONE AND ONLY ONE FUNCTION CALL PER MESSAGE.
\end{lstlisting}

\section{Judge Prompt for Trajectory Quality}
\label{app:judge_prompt}

In our rollout quality analysis (Section~\ref{sec:analysis}), we employ Gemini as a judge to compare trajectories based on efficiency, logic, precision, and safety. The full system prompt used for this evaluation is presented in Listing~\ref{lst:judge_prompt}.

\begin{lstlisting}[
    caption={System prompt for the Gemini judge used in pairwise quality comparison.}, 
    label={lst:judge_prompt},
    basicstyle=\ttfamily\small, % 建议设置字体大小，防止溢出
    breaklines=true,            % 建议开启自动换行
    frame=single                % 建议加上边框使其更清晰
][t]
You are an expert Judge in a Software Engineering competition.
You will be presented with a Problem Statement and TWO different successful trajectories (Trajectory A and Trajectory B) that solved the problem.

Your goal is to decide **WHICH TRAJECTORY IS BETTER**.

Criteria for "Better":
1. **Efficiency**: Fewer wasted steps, less unrelated context reading.
2. **Logic**: Clearer reasoning 
3. **Precision**: No hallucinations, no syntax errors in commands, no reverting edits.
4. **Safety**: Did not delete unrelated files or ruin the environment.

### Problem Statement
{problem}

---
### Trajectory A
{traj_a}

---
### Trajectory B
{traj_b}

---
### Decision
Compare the two objectively.
Return a JSON object with the following structure (Put 'winner' FIRST):
{{
    "winner": "A", "B", or "Tie",
    "reason": "Brief explanation...",
    "better_points": ["Point 1", "Point 2"]
}}
\end{lstlisting}

\end{document}